\documentclass{IEEEtran}

\usepackage[colorlinks,urlcolor=blue,linkcolor=blue,citecolor=blue]{hyperref}

\usepackage{color,array}
\usepackage{graphicx}
\usepackage{academicons}
\usepackage{amsmath}
\usepackage{xcolor}
\usepackage{longtable}
\usepackage{caption}
\usepackage{amsmath,amsfonts}
\usepackage{array}
\usepackage{url}
\usepackage{verbatim}
\usepackage{graphicx}
\usepackage{hyperref}
\usepackage{academicons}
\usepackage{amsmath}
\usepackage{xcolor}
\usepackage{longtable}
\usepackage{caption}
\usepackage{academicons}
\usepackage{caption}
\usepackage{multirow}
\usepackage{adjustbox}
\usepackage{rotating}
\usepackage{pgfplots}
\usepackage{subcaption}
\usepackage{neuralnetwork}
\usepackage{tabularx}
\usepackage{graphicx,subcaption}
\usepackage{lipsum}
\usepackage{algorithm}
\usepackage{algpseudocode}

\newtheorem{lemma}{Lemma}
\begin{document}


\title{Integrating Quantum-Classical Attention in Patch Transformers for Enhanced Time Series Forecasting} 

\author{Sanjay Chakraborty,\thanks{Sanjay Chakraborty is working in the Department of Computer and Information Science (IDA), REAL, AIICS,  at Linköping University, Sweden and Department of Computer Science \& Engineering, Techno International New Town, Kolkata, India (Email: sanjay.chakraborty@liu.se).}
\and Fredrik Heintz\thanks{Fredrik Heintz is working in the Department of Computer and Information Science (IDA), REAL, AIICS, at Linköping University, Sweden (Email: fredrik.heintz@liu.se).}}



\maketitle

\begin{abstract}
QCAAPatchTF is a quantum attention network integrated with an advanced patch-based transformer, designed for multivariate time series forecasting, classification, and anomaly detection. Leveraging quantum superpositions, entanglement, and variational quantum eigensolver principles, the model introduces a quantum-classical hybrid self-attention mechanism to capture multivariate correlations across time points. For multivariate long-term time series, the quantum self-attention mechanism can reduce computational complexity while maintaining temporal relationships. It then applies the quantum-classical hybrid self-attention mechanism alongside a feed-forward network in the encoder stage of the advanced patch-based transformer. While the feed-forward network learns nonlinear representations for each variable frame, the quantum self-attention mechanism processes individual series to enhance multivariate relationships. The advanced patch-based transformer computes the optimized patch length by dividing the sequence length into a fixed number of patches instead of using an arbitrary set of values. The stride is then set to half of the patch length to ensure efficient overlapping representations while maintaining temporal continuity. QCAAPatchTF achieves state-of-the-art performance in both long-term and short-term forecasting, classification, and anomaly detection tasks, demonstrating state-of-the-art accuracy and efficiency on complex real-world datasets. 
\end{abstract}




\section{INTRODUCTION}
\begin{figure}[hbt!]
\begin{center}
\includegraphics[scale=0.25]{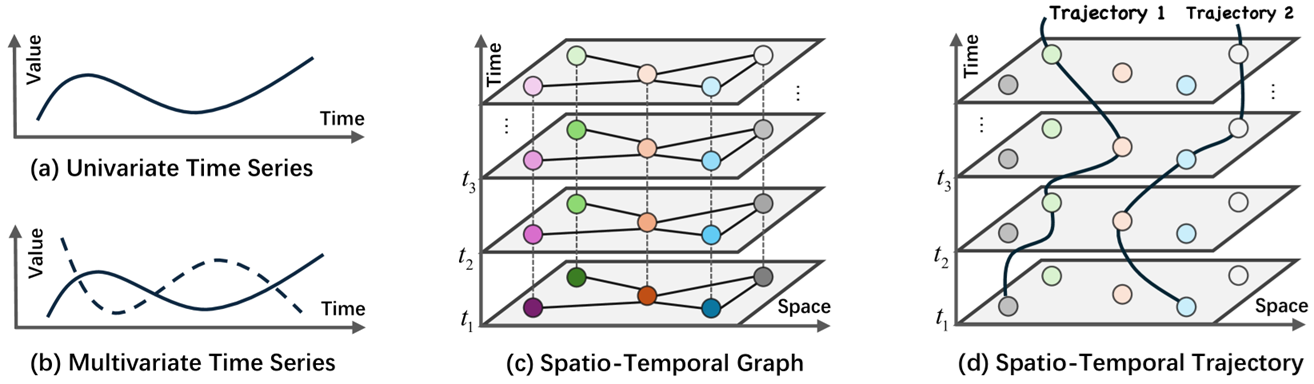}
\caption{Types of Time Series}
\label{TS}
\end{center}
\end{figure}

Time series analysis is a crucial technique in data science, enabling insights into temporal data patterns. It encompasses forecasting, which predicts future values based on historical trends, aiding applications like stock market prediction \cite{zhang2024hybrid}, land-use monitoring \cite{shi2025digital}, energy consumption \cite{yaprakdal2023multivariate}, and weather forecasting \cite{engel2024transformer}. Classification involves categorizing time series data, useful in activity recognition and medical diagnosis \cite{kong2025deep}. Anomaly detection identifies deviations from expected behaviour, essential for fraud detection, industrial fault detection, and network security \cite{xu2021anomaly}. It has also been instrumental in epidemiology and healthcare research \cite{wen2022transformers, morid2023time}. Accurate forecasting is essential for data-driven decision-making in these domains. A notable example is the COVID-19 pandemic (SARS-CoV-2), which, due to its high contagion rate, placed immense strain on healthcare systems worldwide \cite{cajachagua2025impact}. Time series can be classified as either univariate or multivariate, describing one or more variables that change over time, respectively \cite{zhang2024self}. There are two other types of time series. Spatio-temporal trajectory time series captures the movement of objects over time, represented as sequences of spatial coordinates \( (x_k, y_k, z_k) \) with timestamps \( t_k \) and optional contextual features \( f_k \). Formally, a trajectory is defined as, $T_i = \{ (t_k, x_k, y_k, z_k, f_k) \}_{k=1}^{N}$, where \( N \) is the number of time steps \cite{li2023transferable}. A \textit{spatio-temporal graph} (STG) represents dynamic relationships among entities evolving over time. It is defined as \( G = (V, E, X) \), where \( V \) is the set of nodes, \( E \) is the set of edges representing spatial or temporal dependencies, and \( X = \{ X_t \}_{t=1}^{T} \) denotes node features over \( T \) time steps. The adjacency matrix \( A \) encodes spatial relationships, while temporal dependencies are captured via recurrent or attention-based mechanisms, $H_t = f(H_{t-1}, A, X_t; \theta)$, where \( H_t \) represents the node embeddings at time \( t \) and \( f \) is a graph learning function parameterized by \( \theta \) \cite{spadon2021pay}. The various types of time series data are illustrated in Figure \ref{TS}.

Artificial neural networks, which have a non-linear functioning that allows them to outperform classical algorithms, are the foundation of recent time series approaches \cite{spadon2021pay}. Quantum machine learning (QML) has advanced significantly in recent years \cite{biamonte2017quantum}, \cite{cerezo2022challenges}, \cite{peral2024systematic}. In order to improve machine learning algorithms and perhaps speed up difficult computations, quantum machine learning (QML) makes use of quantum computing concepts like superposition and entanglement. Applications of QML algorithms, such as quantum support vector machines, quantum neural networks, and quantum variational classifiers \cite{don2024fusion}, are being investigated for use in drug discovery, materials research, optimization, forecasting, and cryptography \cite{gohel2024quantum, biamonte2017quantum}. QML speeds up drug modeling and medical imaging in the healthcare industry and helps with risk analysis and portfolio optimization in the financial sector. QML is a crucial area in AI research since it has the potential to solve NP-hard problems more effectively than conventional techniques as quantum hardware develops \cite{cerezo2022challenges}. The introduction of transformers that use quantum self-attention mechanisms has revolutionized the processing of time-series data \cite{shi2024qsan}, \cite{wu2023quantum}. Time series analysis is one of the challenging jobs that Quantum Machine Learning (QML) aims to improve efficiency by combining quantum computing with classical machine learning. Compared to conventional models, QML can handle sequential data more effectively by utilizing quantum parallelism, entanglement, and superposition. Applications include quantum variational circuits for time-series forecasting \cite{hsu2024quantum}, quantum recurrent neural networks (QRNNs) and 'Quantum Kernel-Based Long Short-term Memory (QKLSTM)' for financial forecasting and energy demand prediction \cite{thakkar2024improved}, and quantum kernel methods \cite{zhao2024qksan} for better pattern recognition in time-dependent data like medical diagnostics and climate modeling. These developments imply that QML may be able to perform better than traditional methods when dealing with high-dimensional, large-scale time-series data. A basic machine learning operator called the self-attention mechanism (SAM) creates attention ratings straight from individual sequences to make computation easier. SAM was first presented in the Transformer framework and tackles the problem of long-range dependencies that was a problem for previous neural networks, including recurrent neural networks (RNNs) \cite{vaswani2017attention}. According to experimental findings, SAM improves model performance by reducing dependence on outside data while successfully capturing the inherent correlations between features \cite{vaswani2017attention}.

By incorporating the quantum-classical self-attention (QCSA) mechanism into an advanced patch-based transformer for time-series analysis, this work seeks to go beyond conventional full-attention transformers. The main objective is to allow the model to independently strike the best possible balance between forecast accuracy and computational efficiency. This method improves the model's ability to capture complex temporal correlations by incorporating quantum concepts into the self-attention framework. The following are this paper's primary contributions:\\
1. We have introduced a quantum-classical self-attention network (QCSAN) for a proposed advanced patch-based transformer model and described its working procedure for multivariate time series analysis. QCSAN mainly uses three quantum principles (quantum superpositions, quantum entanglement, and variational quantum eigensolver (VQE)) and a quantum-classical hybrid strategy to compute the attention score of the network. \\
2. The proposed advanced patch-based architecture is inspired by the PatchTST \cite{nie2022time}. In the embedding phase, it utilizes an advanced patch embedding with an optimized patch length and stride, systematically evaluated to restructure input data efficiently. The encoder utilizes a hybrid self-attention mechanism to capture temporal dependencies, making it an encoder-based model. A key distinction of this approach is the integration of a hybrid quantum-attention and full-attention module within the encoder layer. The notable differences between the proposed QCAAPatchTF model and PatchTST, in terms of key features, are detailed in Table \ref{summary_comp}.\\
3. We have performed extensive analyses on various time-series data sets. The usefulness of QCAAPatchTF is demonstrated experimentally, where it achieves a significant performance improvement over a set of benchmark models in forecasting, classification, and anomaly detection tasks. Our thorough examination of its architectural choices and embedded modules reveals exciting possibilities and opens the door for more advancements in this field. 

\begin{table*}[hbt!]
\scriptsize
\tiny
\begin{center}
\caption{Summary of Differences among QCAAPatchTF and other state-of-the-art (SOTA) Time Series Transformer models}
\begin{tabular}{|p{1.5cm}|p{3cm}|p{2cm}|p{2cm}|p{2cm}|p{2cm}|p{2.5cm}|}
\hline
Feature             & QCAAPatchTF                                                                         & PatchTST                                                                 & iTransformer                                                            & Informer                                                & Autoformer                                                        & Crossformer                                                          \\ \hline
Patch Embedding     & Advanced patch embedding with optimized and dynamic patch length and stride.        & Standard patch embedding with fixed patch length and stride.             & Uses a learnable embedding with instance normalization.                 & No patching; uses tokenized representations.            & Employs decomposition-based embedding.                            & Uses local and global cross attention for feature extraction.        \\ \hline
Attention Mechanism & Alternates between Quantum Attention (even layers) and Full Attention (odd layers). & Uses Full Attention throughout the model.                                & Integrates instance-wise attention for adaptive feature weighting.      & Uses a ProbSparse Self-Attention for efficiency.        & Introduces Auto-Correlation Attention for long-term dependencies. & Applies cross attention to capture hierarchical dependencies.        \\ \hline
Normalization       & Normalization and de-normalization of the input/output time series.                 & May include normalization but lacks the custom de-normalization process. & Instance normalization to stabilize learning.                           & Uses standard layer normalization.                      & Combines normalization with trend-seasonality decomposition.      & Normalization is applied per sub-series block.                       \\ \hline
Task-Specific Head  & Dynamically adjusts for forecasting, anomaly detection, or classification tasks.    & Static heads based on the task.                                          & Uses a task-specific MLP-based decoder.                                 & Specialized decoder for long-sequence forecasting.      & Decomposes series into trend and seasonality before decoding.     & Adopts cross-attention-based reconstruction.                         \\ \hline
Efficiency          & Optimized for balanced performance and accuracy.                                    & Heavy memory usage due to full attention.                                & Efficient due to instance normalization and adaptive feature weighting. & Significantly reduces complexity with sparse attention. & Reduces computation by focusing on periodic patterns.             & Balances computational efficiency and accuracy with cross attention. \\ \hline
\end{tabular}
\label{summary_comp}
\end{center}
\end{table*}

\section{BACKGROUND}
\label{background}
\subsection{Transformers for Time Series}
Transformers have gained significant attention in both short-term and long-term forecasting due to their ability to capture complex temporal dependencies \cite{kitaev2020reformer}, \cite{ zeng2023transformers}, \cite{zerveas2021transformer}. Among the early advancements, Informer \cite{zhou2021informer} introduced a generative-based decoder and 'Probability-Sparse' self-attention to address the challenge of quadratic time complexity. Building on this, models such as Autoformer \cite{chen2021autoformer}, iTransformer \cite{liu2023itransformer}, FEDFormer \cite{zhou2022fedformer}, PatchTST \cite{nie2022time}, ETSformer \cite{woo2022etsformer}, and EDformer \cite{chakraborty2024edformer} have further enhanced time-series modeling. iTransformer \cite{liu2023itransformer} innovates by representing individual time points as variate tokens, enabling the attention mechanism to model multivariate correlations while leveraging feed-forward networks to learn nonlinear representations. PatchTST \cite{nie2022time} enhances local and global dependency capture through patch-based processing, while Crossformer \cite{zhang2023crossformer} introduces a dimension-segment-wise (DSW) technique that encodes time-series data into a structured 2D representation. The core strength of transformer models lies in their attention mechanism, which allows them to focus on critical segments of the input sequence for accurate predictions \cite{zeng2023transformers}. By computing weighted representations through attention scores between query, key, and value vectors, transformers effectively capture dependencies across sequence positions, regardless of their distance. Scaled dot-product attention ensures stable gradient propagation, while multi-head attention extends this capability by learning diverse patterns from different input subspaces. 

\subsection{Quantum Logic}
Quantum computing leverages quantum mechanics principles such as superposition, entanglement, teleportation, and quantum interference to process information exponentially faster than classical computers for certain tasks. Instead of classical bits (0 or 1), quantum bits (qubits) exist in a superposition of both states simultaneously, enabling parallel computations. Quantum gates manipulate qubits through unitary transformations, enabling powerful algorithms like Shor’s for factorization and Grover’s for search optimization, quantum variational algorithms, and AI advancements \cite{biamonte2017quantum}.

\subsubsection{Quantum Superposition}
Quantum superposition states that a qubit can exist in a linear combination of both \( |0\rangle \) and \( |1\rangle \) states simultaneously. Mathematically, this is represented as:
\[
|\psi\rangle = \alpha |0\rangle + \beta |1\rangle
\]
where \( \alpha \) and \( \beta \) are complex probability amplitudes satisfying:
\[
|\alpha|^2 + |\beta|^2 = 1
\]
Upon measurement, the qubit collapses to \( |0\rangle \) with probability \( |\alpha|^2 \) or \( |1\rangle \) with probability \( |\beta|^2 \). This enables quantum computers to explore multiple states in parallel, offering significant computational advantages over classical systems  \cite{wu2023quantum} \cite{hsu2024quantum}.

\subsubsection{Quantum Entanglement}
Quantum entanglement is a phenomenon where two or more qubits become correlated in such a way that the state of one qubit is instantly dependent on the state of the other, regardless of the distance between them. A common example is the Bell state:
\[
|\Phi^+\rangle = \frac{1}{\sqrt{2}} (|00\rangle + |11\rangle)
\]
Here, the two qubits exist in a superposition of both \( |00\rangle \) and \( |11\rangle \). Measuring one qubit immediately determines the state of the other, demonstrating non-local correlations. This property is fundamental to quantum communication, cryptography, and computing  \cite{wu2023quantum} \cite{hsu2024quantum}.

\subsection{Variational Quantum Algorithms}
The Variational Quantum Eigensolver (VQE) and Variational Quantum Classifier (VQC) are two hybrid quantum-classical algorithms leveraging parameterized quantum circuits for optimization and machine learning tasks \cite{don2024fusion}. In VQE, a quantum subroutine is run inside of a classical optimization loop \cite{zhang2025diffusion}. VQE is used to find the ground-state energy of a given Hamiltonian \( H \) by minimizing the expectation value of the Hamiltonian over a parameterized quantum state \( |\psi(\theta)\rangle \):
\[
E(\theta) = \langle \psi(\theta) | H | \psi(\theta) \rangle
\]
The parameters \( \theta \) are optimized using a classical optimizer, such as gradient descent, to iteratively refine the quantum state. This method is crucial for quantum chemistry and materials science.

\begin{figure}[hbt!]
\begin{center}
\includegraphics[scale=0.4]{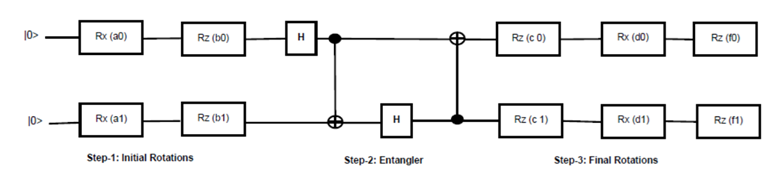}
\caption{Variational Quantum Circuit}
\label{VQE}
\end{center}
\end{figure}

VQC applies a similar variational approach to quantum machine learning \cite{maheshwari2021variational}. Given an input data point \( x \), it is encoded into a quantum state \( |\psi(x)\rangle \), which is processed through a parameterized quantum circuit \( U(\theta) \):
\[
|\phi(x, \theta)\rangle = U(\theta) |\psi(x)\rangle
\]
A measurement operator \( M \) is then applied to extract the classification decision:
\[
y = \langle \phi(x, \theta) | M | \phi(x, \theta) \rangle
\]
The parameters \( \theta \) are trained using a classical optimizer to minimize a loss function, enabling quantum-enhanced classification. Both VQE and VQC demonstrate the power of variational quantum algorithms, balancing quantum computation with classical optimization to solve multi-class complex problems efficiently \cite{zhou2023multi}. In Figure \ref{VQE}, the 'Variational Quantum Classifier (VQC)' circuit consists of three key stages: initial rotations, entanglers, and final rotations. Initially, RX, RY, or RZ gates encode classical data into quantum states. Next, entangling layers, typically using CNOT (CX) gates, create quantum correlations between qubits. 

Finally, trainable rotation gates refine the quantum state before measurement. The circuit parameters are optimized using classical techniques to minimize a loss function, enabling effective quantum classification.

\section{PROBLEM STATEMENTS}
\label{prob}
This work deals with the challenges of long-term and short-term multivariate time series (MTS) forecasting by utilizing historical data while also considering classification and anomaly detection tasks. A MTS at time t is defined as $({X}_t = [x_{t,1}, x_{t,2}, \dots, x_{t,N}])$, where $x_{t,n}$ denotes the value of the n-th variable at time t for n = 1, 2,...., N. The notation ${X}_{t:t+H}$ is used to represent the series values from time t to t+H, inclusive. However, for a given starting time $t_0$, the model receives as input the sequence ${X}_{t_0-L:t_0}$, representing the last L time steps, and produces the predicted sequence \( \hat{\mathbf{X}}_{t_0:t_0+H} \), corresponding to the forecasted values for the following \( H \) time steps. The forecasted value at any time t is denoted as \( \hat{\mathbf{X}}_t \). 
\begin{equation}
\hat{\mathbf{X}}_{t:t+H} = f(\mathbf{X}_{t-L:t})
\end{equation}
Given a time series dataset \( X = \{ X_1, X_2, \dots, X_N \} \), where \( X_i = [ x_{i,1}, x_{i,2}, \dots, x_{i,T} ] \) represents a sequence of observations, the objective is to assign each sequence \( X_i \) to one of \( C \) possible classes \( \{ y_1, y_2, \dots, y_C \} \). The challenge lies in capturing both global and local temporal dependencies, handling noisy and irregular data, and ensuring robustness across various time series lengths. Applications span diverse domains, including healthcare, finance, and activity recognition, where accurate classification is critical for decision-making \cite{campagner2024ensemble}.
\begin{equation}
y_i = \arg\max_{c \in \{y_1, y_2, \dots, y_C\}} P(y = c \mid X_i)
\end{equation}
where \( P(y = c \mid X_i) \) is the probability of class \( c \) given the input time series \( X_i \), and the objective is to assign the label \( y_i \) to the class with the highest probability.

Time series anomaly detection aims to find irregular patterns within temporal data that deviate from normal behaviour. Given a time series \( X = [ x_1, x_2, \dots, x_T ] \), the goal is to detect instances \( t \) where \( x_t \) or a segment \( X_{t:t+k} \) exhibits anomalies. These anomalies may arise due to faults, unusual events, or rare occurrences, and their detection is crucial in applications such as system monitoring, fraud detection, and predictive maintenance. The task is complicated by the need to distinguish genuine anomalies from noise, adapt to non-stationary data, and minimize false positives while ensuring timely detection.
\begin{equation}
\mathcal{A} = \{ t : |x_t - \hat{x}_t| > \epsilon \}
\end{equation}
where \( \hat{x}_t \) is the predicted value of \( x_t \) based on past observations, and \( \epsilon \) is a predefined threshold that determines if the deviation is considered an anomaly.

\section{METHODOLOGY}
\label{Method}
The QCAAPatchTF model is designed for time series forecasting, anomaly detection, and classification, integrating both classical and quantum attention mechanisms. The overall algorithm of the proposed QCAAPatchTF approach for all three tasks (forecasting, classification, and anomaly detection) is described in Algorithm \ref{algo2}.

\begin{figure*}[hbt!]
\begin{center}
\includegraphics[scale=0.25]{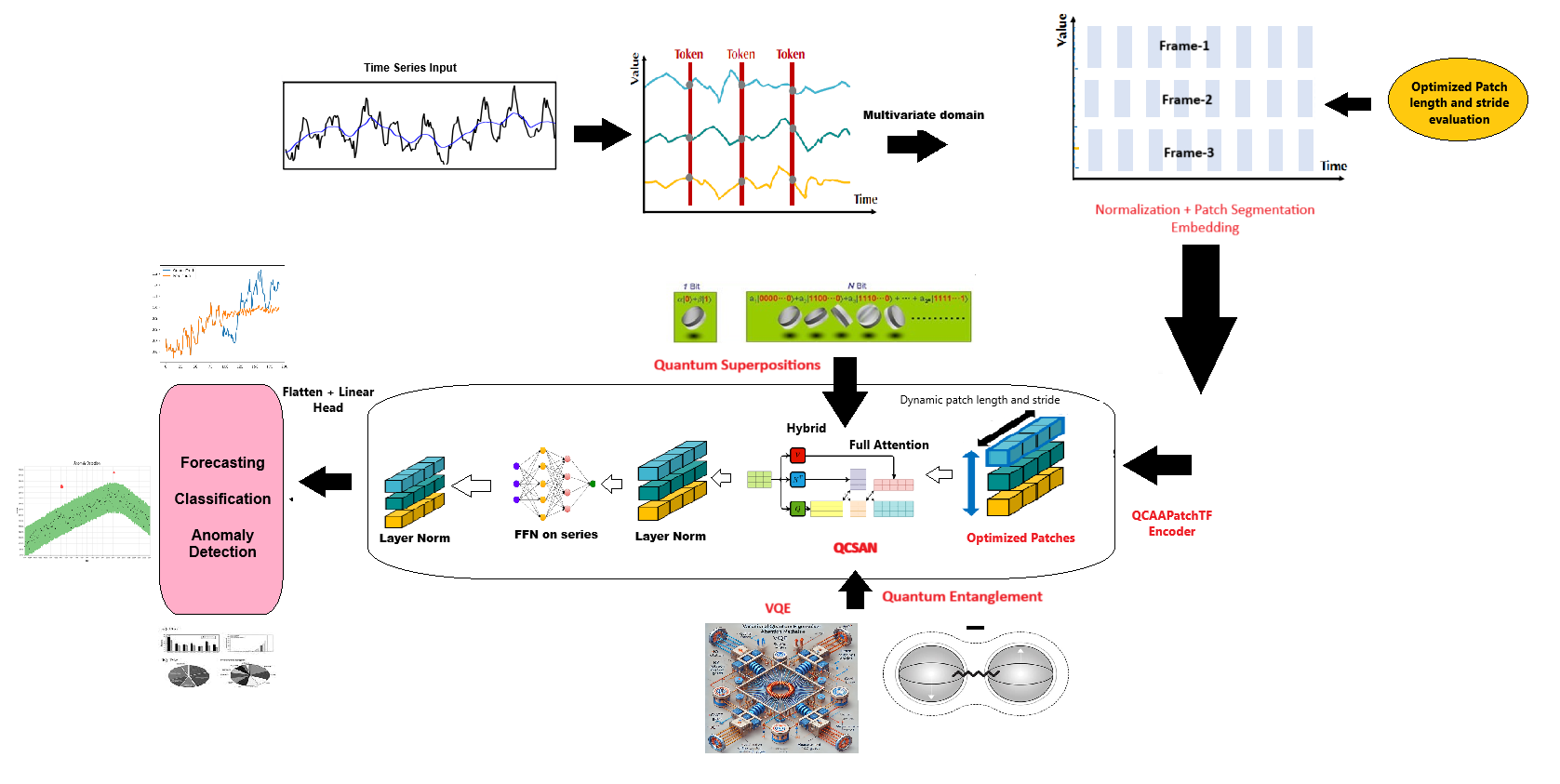}
\caption{Overall approach of QCAAPatchTF}
\label{QCAAPatchTF_architecture}
\end{center}
\end{figure*}

\subsection{Model Inputs}
\begin{figure}[hbt!]
\begin{center}
\includegraphics[scale=0.25]{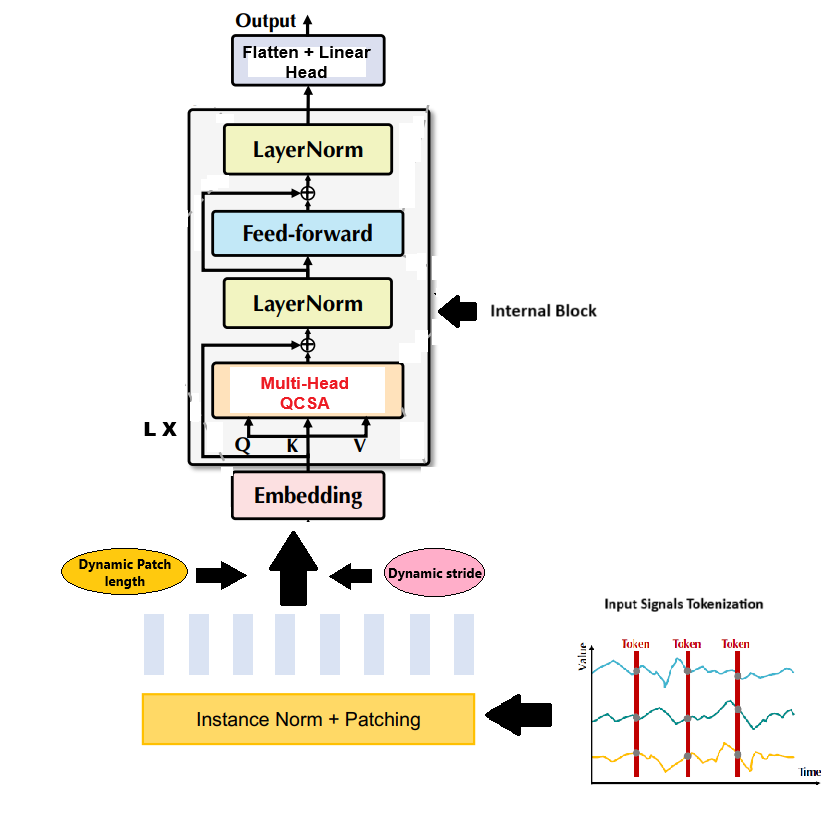}
\caption{Architecture of QCAAPatchTF internal blocks}
\label{Fig2}
\end{center}
\end{figure}

Our proposed encoder-only QCAAPatchTF design encourages adaptive correlation and representation learning in multivariate series. The unique features of each component are captured by tokenizing each time series into a set of patches. This design captures complex temporal dependencies in time series well. Let 'Pl' denote the patch length and 'St' represents the stride, the non-overlapping region between two consecutive patches. Unlike the traditional PatchTST approach that uses fixed or arbitrary patch lengths and strides, this model dynamically computes an optimized patch length (OPl) based on the sequence length (\(\text{seq\_len}\)). The ``evaluate'' method ensures a structured approach to determine the number of patches (defaulting to 6 if not specified) and calculates the patch length as:
\[
\text{OPl} = \frac{\text{seq\_len}}{\text{num\_patches}}
\]
To ensure overlapping patch embeddings, which help retain temporal dependencies, the optimized stride (OSt) is then set to half of the computed patch length:
\[
\text{OSt} = \frac{\text{patch\_len}}{2}
\]
Table \ref{evalOplOst} presents an analysis of optimized patch lengths and strides for different sequence lengths across various time series tasks, including long-term forecasting, short-term forecasting, anomaly detection, and classification. It highlights how different sequence:2.1770832538604736, mae:1.1801297664642334, dtw:-9 lengths require varying patch and stride configurations to capture temporal dependencies effectively. Additionally, padding is initialized to match the stride value, ensuring proper alignment and preserving critical sequence information. This approach avoids arbitrary choices and enhances learning efficiency, leading to improved feature representation in downstream forecasting or classification tasks.
\begin{equation}
\begin{split}
h^0_n=Embedding(Patches(X:,n), OPl, OSt)\\
H^{(l+1)}=IntBlock(H^l), l= 0, ....., L-1, \\
Y^t:,n = Projection(h^L_n),
\end{split}
\end{equation}
Where the superscript denotes the layer index, and \( H = \{ h_1, \dots, h_N \} \in \mathbb{R}^{N \times D} \) consists of \( N \) embedded tokens of size \( D \). Multi-layer perceptrons (MLPs) are used for projection. 
\begin{table}[hbt!]
\scriptsize
\tiny
\begin{center}
\caption{Analysis of optimized patch lengths and strides across various sequence lengths for different tasks. The red-colored values indicate the cases used in this study.}
\begin{tabular}{|c|c|c|c|}
\hline
Tasks                                                                               & Seq\_len                   & Patch\_len                & Stride                   \\ \hline
                                                                                    & {\color[HTML]{FE0000} 96}  & {\color[HTML]{FE0000} 16} & {\color[HTML]{FE0000} 8} \\ \cline{2-4} 
                                                                                    & 240                        & 40                        & 20                       \\ \cline{2-4} 
\multirow{-3}{*}{\begin{tabular}[c]{@{}c@{}}Long-term \\ Forecasting\end{tabular}}  & 420                        & 70                        & 35                       \\ \hline
                                                                                    & 24                         & 4                         & 2                        \\ \cline{2-4} 
                                                                                    & 48                         & 8                         & 4                        \\ \cline{2-4} 
\multirow{-3}{*}{\begin{tabular}[c]{@{}c@{}}Short-term \\ Forecasting\end{tabular}} & {\color[HTML]{FE0000} 96}  & {\color[HTML]{FE0000} 16} & {\color[HTML]{FE0000} 8} \\ \hline
Anomaly                                                                             & {\color[HTML]{FE0000} 100} & {\color[HTML]{FE0000} 17} & {\color[HTML]{FE0000} 8} \\ \hline
Classification                                                                      & 512                        & 85                        & 41                       \\ \hline
\end{tabular}
\label{evalOplOst}
\end{center}
\end{table}
By transforming input signals into patches, we strengthen local dependencies and capture rich semantic information by grouping time steps into subseries-level patches. Patching on \textit{time series signals} involves segmenting the input sequence into patches, to enhance temporal locality and feature extraction. Given a univariate time series \( x^{(i)} \) of length \( L \), we divide it into optimized patches of length \( OPl \) with an evaluated optimized stride \( OSt \), generating a sequence of patches \( x^{(i)}_p \in \mathbb{R}^{OPl \times N} \), where 
\[
N = \left\lfloor \frac{L - OPl}{OSt} \right\rfloor + 2.
\]
To preserve the sequence length, the last value \( x^{(i)}_L \in \mathbb{R} \) is padded \( OSt \) times at the end before patching. This transformation reduces the number of input tokens from \( L \) to approximately \( L/S \), significantly lowering the attention map’s memory usage and computational complexity by a factor of \( S \). Consequently, patching enables the model to process longer historical sequences, improving forecasting performance while optimizing training time and GPU memory.
The \texttt{IntBlock()} processes each frame individually via a shared feed-forward network, with interactions facilitated by quantum classical self-attention. The internal architecture of QCAAPatchTF is illustrated in Figure \ref{Fig2}.

\subsection{Encoding of Model}
In this block, we have organized a stack of 'L' number of blocks, each consisting of the proposed quantum classical attention network (QCAN), feed-forward network, and layer normalization modules.

\subsubsection{Quantum Classical Self-Attention (QCSA)}
The quantum-classical self-attention mechanism is extensively employed to model temporal dependencies in forecasting, classification, and anomaly detection. This approach enables dynamic weighting of sequence tokens, effectively capturing both long-range and short-range dependencies along with intricate contextual relationships. In classical attention models, given input queries Q, keys K, and values V, attention scores are computed using the dot product of query and key vectors, followed by normalization and weighted summation to generate updated embeddings. Our quantum-classical self-attention extends this process by integrating quantum principles such as superposition, entanglement, and variational quantum algorithms, enhancing the representation of complex dependencies between data points. This hybrid attention module integrates seamlessly into a transformer encoder, making it highly effective for sequence-based applications. This is called hybrid attention as it dynamically switches between quantum attention and full attention for each encoder layer in a transformer model. Quantum attention is applied for even layers and full attention is applied in odd layers. Quantum superposition allows a system to exist in multiple states simultaneously, while entanglement ensures strong correlations between elements, enabling richer and more efficient modeling of sequential relationships. The \textit{QuantumClassicalAttention} module integrates the\textit{ Variational Quantum Eigensolver (VQE)} to compute attention scores, using a \textit{PennyLane} quantum circuit with \textit{RY rotations and CNOT gates} for parameterized encoding and entanglement. Given input queries \( Q \in \mathbb{R}^{B \times L \times H \times E} \) and keys \( K \in \mathbb{R}^{B \times S \times H \times D} \), the attention mechanism first computes superposition scores via tensor contractions:
\[
S = QK^T, \quad S \in \mathbb{R}^{B \times H \times L \times S}
\]
These scores are then processed by the quantum circuit, where each qubit undergoes RY rotations based on learnable parameters \( \theta \):
\[
\left| \psi(\theta) \right\rangle = \bigotimes_{i=1}^{n} RY(\theta_i) \left| 0 \right\rangle
\]
and \textit{CNOT gates} create entanglement:
\[
U_{\text{ent}} = \prod_{i=1}^{n-1} CNOT(i, i+1)
\]
The \textit{quantum attention score} is derived from the expectation value of the \textit{Pauli-Z operator} on the first qubit:
\[
A_q = \langle \psi(\theta) | Z | \psi(\theta) \rangle
\]
Additionally, an \textit{entanglement-aware score} is computed via another tensor contraction:
\[
A_e = VK^T, \quad A_e \in \mathbb{R}^{B \times H \times L \times S}
\]
The final attention score combines quantum and entanglement terms:
\[
A = A_q + \lambda A_e
\]
where \( \lambda \) is a tunable entanglement factor. If a mask \( M \) is applied, we set:
\[
A = A + M \cdot (-\infty)
\]
Softmax normalization follows:
\[
A' = \text{softmax}(A)
\]
which is then used to compute the final \textit{attention-weighted} values:
\[
V' = A'V
\]
This hybrid quantum-classical approach enhances feature learning by leveraging quantum entanglement and quantum variational techniques, making it valuable for time series forecasting, NLP, and anomaly detection. For multi-head superposition-like states, each attention head transforms the input \( Q, K, V \) using different projection matrices \( W_Q^h, W_K^h, W_V^h \), where \( h \) is the head index:

\begin{equation}
Q_h = W_Q^h Q, \quad K_h = W_K^h K, \quad V_h = W_V^h V
\end{equation}

The updated embeddings per head are computed as:

\begin{equation}
E_h = \sigma \left( W_V^h \cdot V' \right)
\end{equation}

\textit{Multi-Head Concatenation:}
All heads’ outputs are concatenated:

\begin{equation}
E_{\text{multi-head}} = \text{Concat}(E_1, E_2, \dots, E_H)
\end{equation}

A final projection matrix \( W_O \) is applied:

\begin{equation}
E_{\text{final}} = W_O E_{\text{multi-head}}
\end{equation}

A residual connection and layer normalization are added to stabilize training.

\begin{equation}
E_{\text{output}} = \text{LayerNorm}(E_{\text{final}} + Q)
\end{equation}

\begin{figure}[hbt!]
\begin{center}
\includegraphics[scale=0.25]{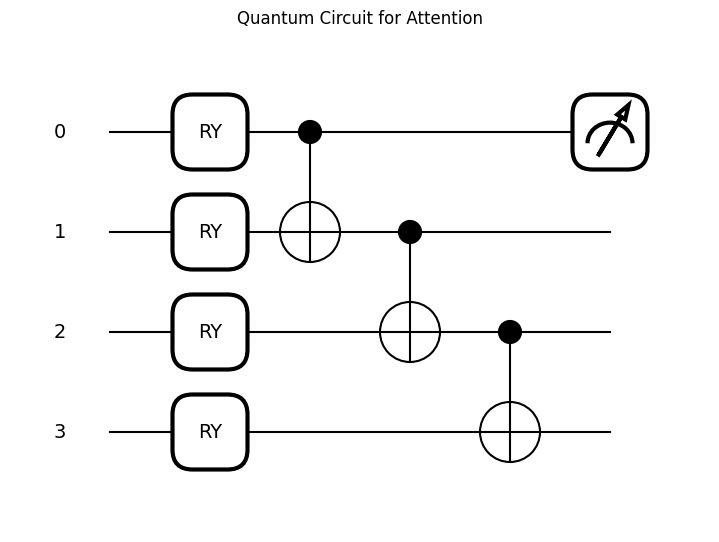}
\caption{VQE-based Quantum Attention Circuit for Each Head}
\label{vqeqsac}
\end{center}
\end{figure}

The quantum-classical self-attention mechanism for a single head is described in Algorithm \ref{algo1}. Figure \ref{vqeqsac} represents the quantum attention circuit design for each head. 

\subsubsection{Normalization of Layers}
The following block, "Layer Normalization (LayerNorm)," is added  to enhance deep networks' training stability and convergence. In most transformer-based forecasters, this module normalizes the multivariate representation of the same timestamp by gradually integrating variables. However, our advanced design normalizes the series representation of individual variates, as shown in Equation \ref{eq11}. 
\begin{equation}
\label{eq11}
LayerNorm(H) = \left[\frac{ h_n - Mean(h_n) } { \sqrt{Var(h_n)} } | n = 1, ....., N \right]
\end{equation}

\subsubsection{Feed-forward network}
The feed-forward network (FFN), which is applied consistently to each patch-based frame in this instance, is the core element of the Transformer for token representation encoding. The universal approximation theorem states that these networks are capable of modeling time series data by capturing intricate representations. They focus on capturing the observed time series and stacking advanced patches to decode the representations for subsequent series utilizing dense non-linear connections. The feed-forward and multi-head attention blocks are iterated n times in the encoder block.
\begin{equation}
FFN(H')=ReLU(H'W^1+b^1)W^2+b^2
\end{equation}
Where, \( H' \) is the output of the previous layer, and \( W_1, W_2, b^2 \) are trainable parameters.

\begin{equation}
H'=LayerNorm(MVSelfAtten(X)+X)
\end{equation}
\begin{equation}
H=LayerNorm(FFN(H')+H')
\end{equation}
Where, \( \text{MVSelfAtten}(\cdot) \) denotes the self-attention module for multivariate and \( \text{LayerNorm}(\cdot) \) defines the layer normalization job. 

\subsubsection{Loss Function}
We have chosen to use the 'Mean Squared Error (MSE)' loss to measure the discrepancy between the prediction and the ground truth. The loss for each channel is computed and then averaged over \( M \) time series to obtain the overall objective loss:

\begin{equation}
\mathcal{L} = \mathbb{E}_{x} \frac{1}{M} \sum_{i=1}^{M} \left\| \hat{x}^{(i)}_{L+1:L+T} - x^{(i)}_{L+1:L+T} \right\|_2^2
\end{equation}

where \( \hat{x}^{(i)}_{L+1:L+T} \) represents the predicted values for time series \( i \), and \( x^{(i)}_{L+1:L+T} \) denotes the corresponding ground truth values over the forecasting horizon \( T \). This loss function ensures that the model minimizes the squared error between predictions and actual values across all time series \cite{nie2022time}.

\subsubsection{Normalization of Instances}
It normalizes each time series instance \( x^{(i)} \) to have zero mean and unit standard deviation. Essentially, we normalize each \( x^{(i)} \) before patching, and then the mean and standard deviation are added back to the output prediction \cite{nie2022time}.

\section{Time Complexity}
The \textit{QuantumClassicalAttention} module has an overall computational complexity $O(B H L S D)$ dominated by tensor contractions for superposition and entanglement-aware scores. The quantum circuit, using $n = O(\log S)$ qubits, incurs an additional cost $O(n)$ for \textit{RY rotations} and $O(n)$ for \textit{CNOT} entanglement, making its contribution logarithmic. The final steps, including masking and softmax, run in $O(B H L S)$. Thus, the quantum component adds minimal overhead, keeping the module primarily constrained by classical tensor operations. So, 'BHLSD' collectively represents the tensor shape for the attention computation. Here, B is the batch size, H is the number of heads, L is the query length, S is the key length, and D is the dimensionality of each vector in the sequence.
The \textit{Quantum-Classical Advanced Patch-based Transformer} involves multiple steps, including advanced-patch embedding, attention, and projection. The advanced-patch embedding operation for input \( X_{\text{enc}} \) runs in \( O(B L d_{\text{model}}) \). The \textit{QuantumAttention} has complexity \( O(B H L S D) \), with a quantum overhead of \( O(\log S) \) for \textit{RY} rotations and \textit{CNOT} gates. The \textit{FullAttention} operates with \( O(B H L S D) \) complexity due to softmax calculations. The encoder processes the embeddings in \( O(B H L d_{\text{model}}) \), while the projection layer incurs \( O(B L d_{\text{model}} P) \). The overall complexity is dominated by the attention mechanism and the encoder, yielding \( O(B H L S D + B L d_{\text{model}} P) \), with the quantum overhead contributing logarithmically.

\section{RESULT ANALYSIS}
\label{result}
\subsection{Datasets Description and Implementation Details}
This paper uses data sets that span long-term and short-term time series forecasting, classification, and anomaly detection tasks. Table \ref{table-dataset} provides a detailed overview of the datasets used in this study. All datasets used in this study are publicly available and are partitioned into training, validation, and test sets within the benchmark Time-Series Library (TSLib) \cite{sa_timeseries}. In addition, the M4 experimental short-term dataset is described in Table \ref{M4dataset}. The details of the datasets are described in the supplementary document.

Table \ref{table-impl} outlines the hyperparameters of the QCAAPatchTF approach tailored for four distinct tasks: long-term forecasting, short-term forecasting, classification, and anomaly detection. Key parameters such as d\_model, channel\_independence, and the number of scales (k) are adjusted to suit the specific requirements of each task, while other settings such as batch size, learning rate, and early stopping patience are optimized to balance performance and computational efficiency \cite{vaswani2017attention}. In particular, the number of epochs varies significantly, with classification requiring the most training iterations (50 epochs), reflecting the complexity of learning high-level representations for this task. These configurations ensure that the model is well adapted to the unique challenges posed by each application domain. The details are described in the supplementary document.

\subsection{Long-term and Short-term Forecasting}
\begin{figure}[hbt!]
  \centering
  \begin{subfigure}{0.3\columnwidth}
    \includegraphics[width=\textwidth]{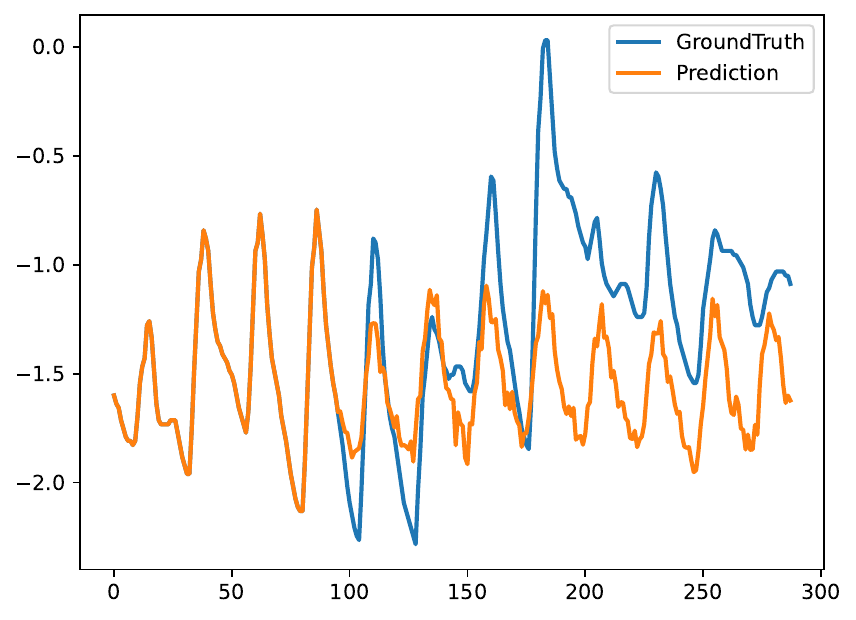}
    \caption{QCAAPatchTF} 
  \end{subfigure}
  \hfill
  \begin{subfigure}{0.3\columnwidth}
    \includegraphics[width=\textwidth]{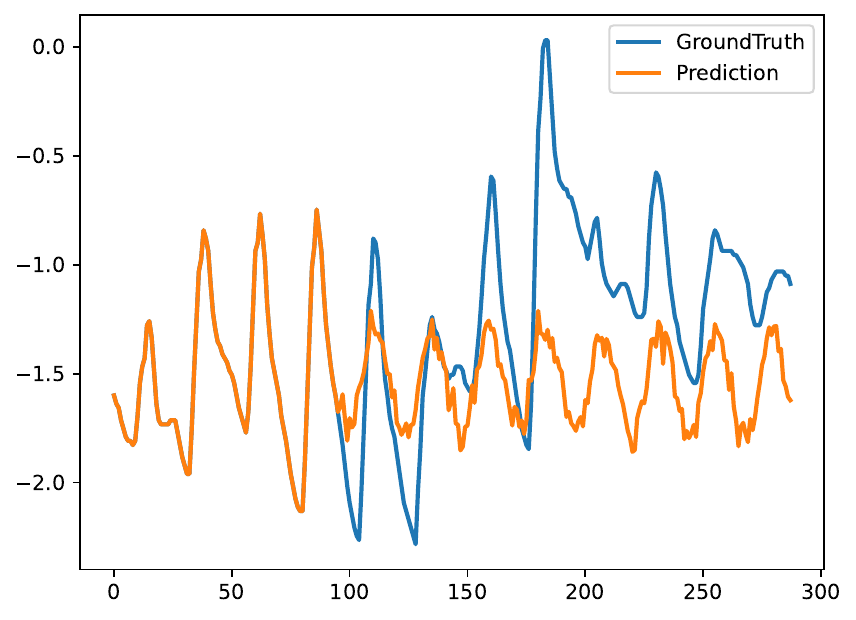}
    \caption{iTransformer} 
  \end{subfigure}
  \hfill
  \begin{subfigure}{0.3\columnwidth}
    \includegraphics[width=\textwidth]{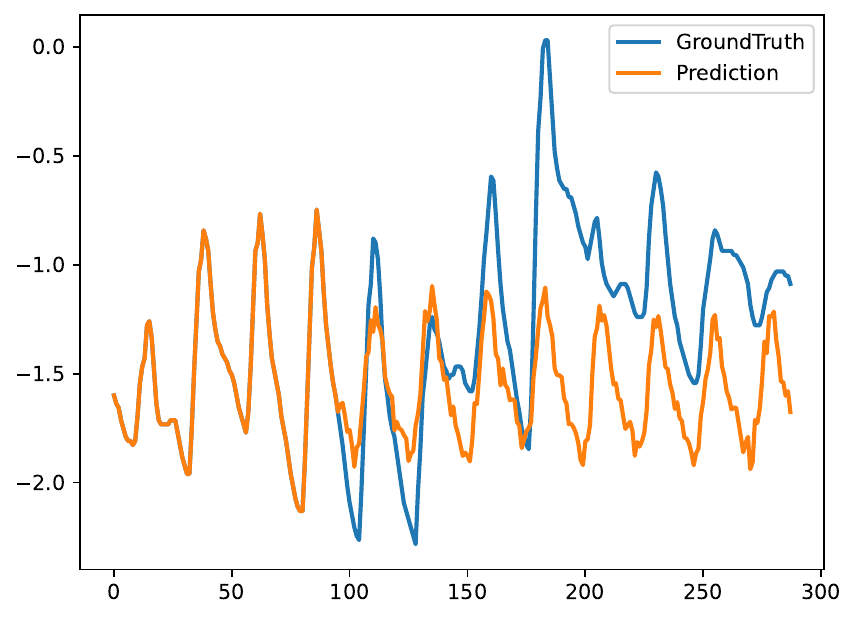}
    \caption{PatchTST} 
  \end{subfigure}

  \begin{subfigure}{0.3\columnwidth}
    \includegraphics[width=\textwidth]{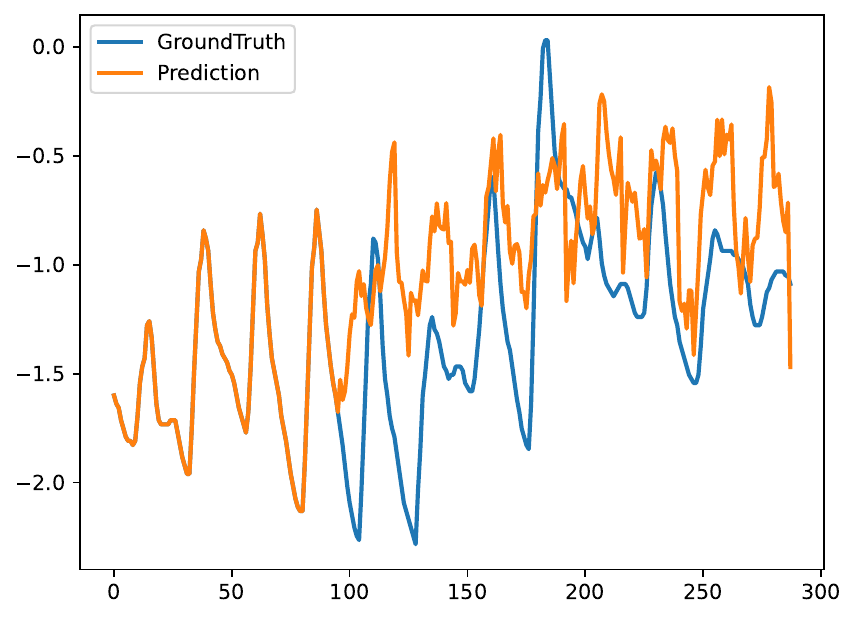}
    \caption{Reformer} 
  \end{subfigure}
  \hfill
  \begin{subfigure}{0.3\columnwidth}
    \includegraphics[width=\textwidth]{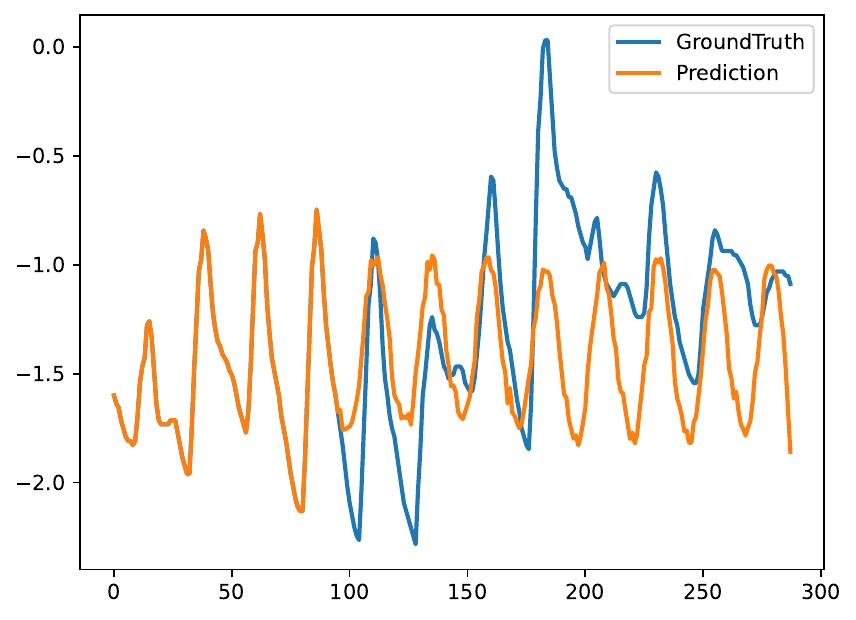}
    \caption{FEDformer} 
  \end{subfigure}
  \hfill
  \begin{subfigure}{0.3\columnwidth}
    \includegraphics[width=\textwidth]{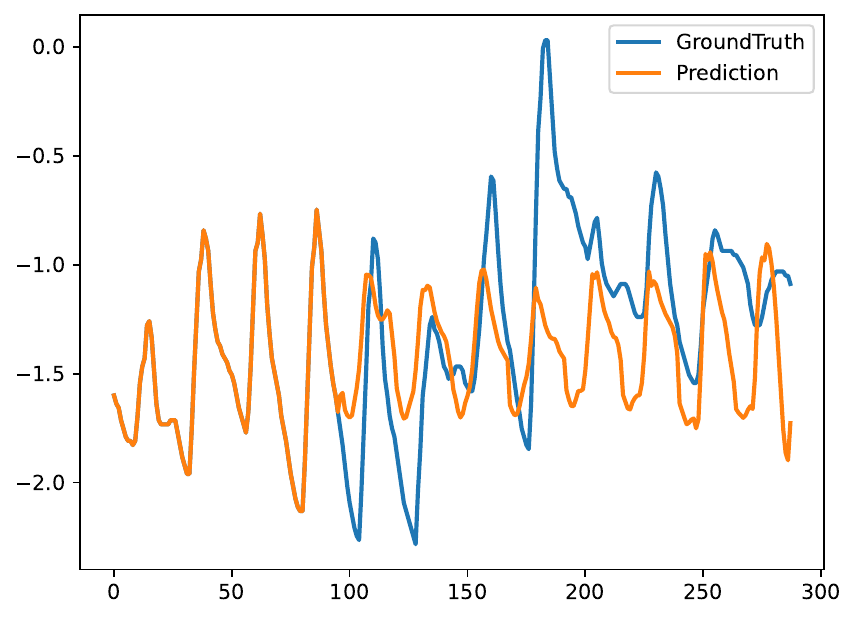}
    \caption{Autoformer} 
  \end{subfigure}

  \begin{subfigure}{0.3\columnwidth}
    \includegraphics[width=\textwidth]{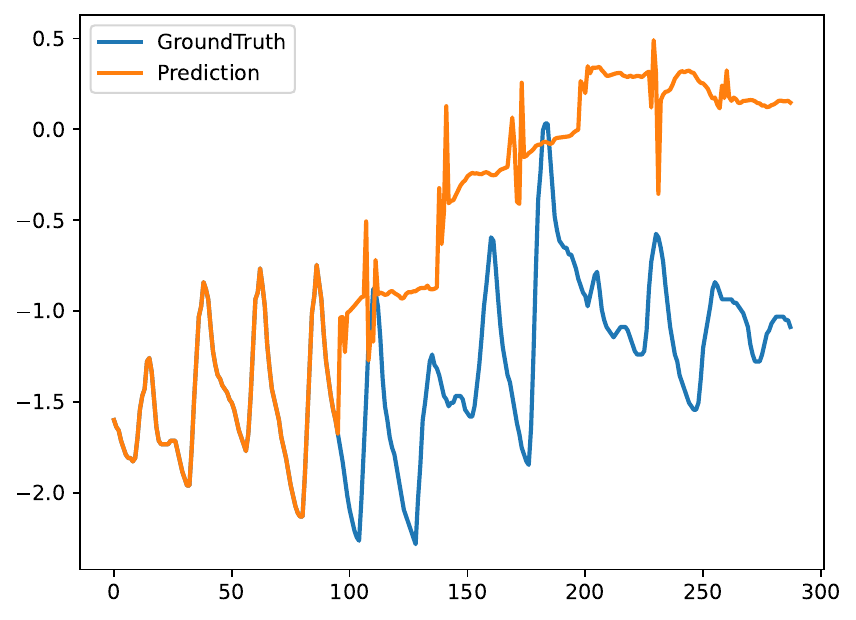}
    \caption{Informer} 
  \end{subfigure}
 \hfill
  \begin{subfigure}{0.3\columnwidth}
    \includegraphics[width=\textwidth]{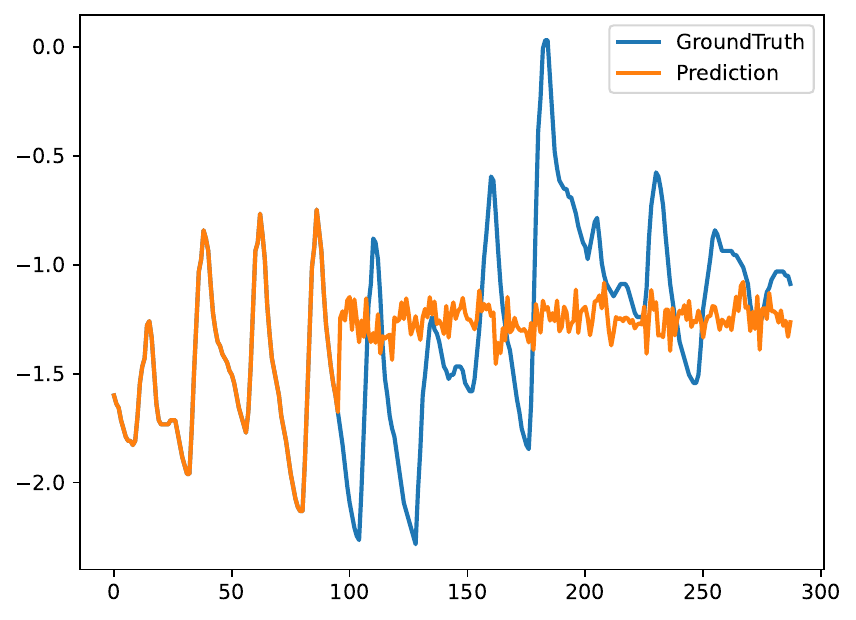}
    \caption{Crossformer} 
  \end{subfigure}
  \hfill
  \begin{subfigure}{0.3\columnwidth}
    \includegraphics[width=\textwidth]{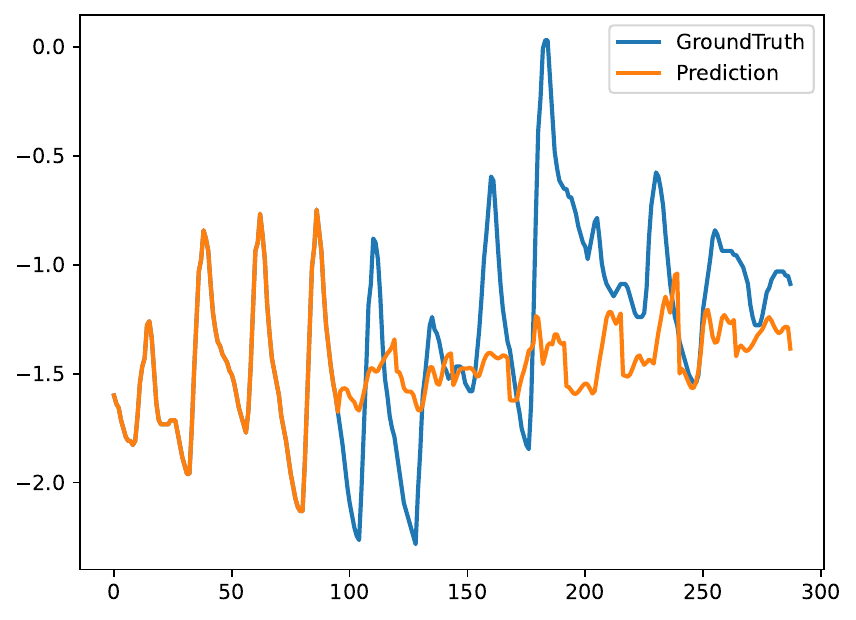}
    \caption{NS-Transformer} 
  \end{subfigure}

  \caption{Visualization of predictions (length:192) on ETTh2 dataset}
  \label{predtestetth2}
\end{figure}

\begin{figure}[hbt!]
  \centering
  \begin{subfigure}{0.3\columnwidth}
    \includegraphics[width=\textwidth]{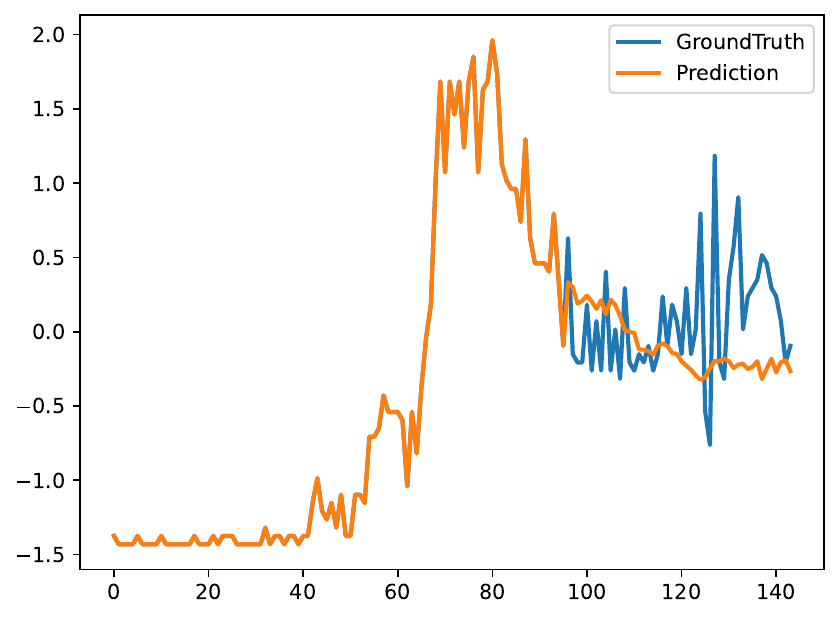}
    \caption{QCAAPatchTF} 
  \end{subfigure}
  \hfill 
  \begin{subfigure}{0.3\columnwidth}
    \includegraphics[width=\textwidth]{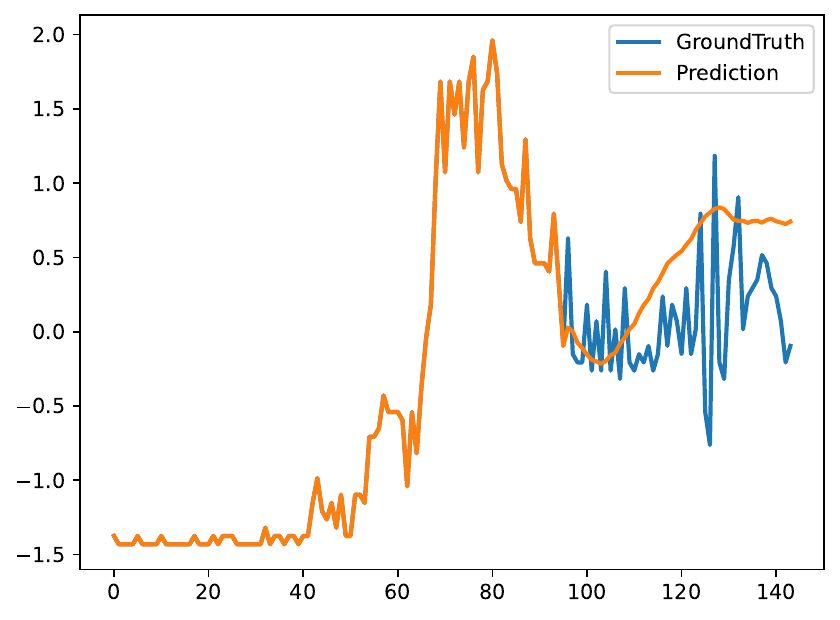}
    \caption{EDformer} 
  \end{subfigure}
  \hfill
  \begin{subfigure}{0.3\columnwidth}
    \includegraphics[width=\textwidth]{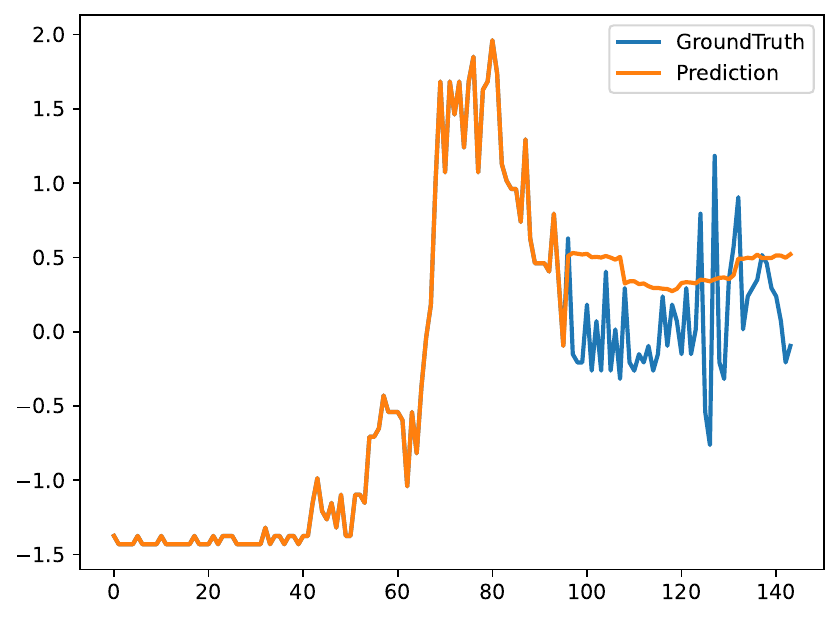}
    \caption{Crossformer} 
  \end{subfigure}
  
  \vspace{1em} 
  \begin{subfigure}{0.3\columnwidth}
    \includegraphics[width=\textwidth]{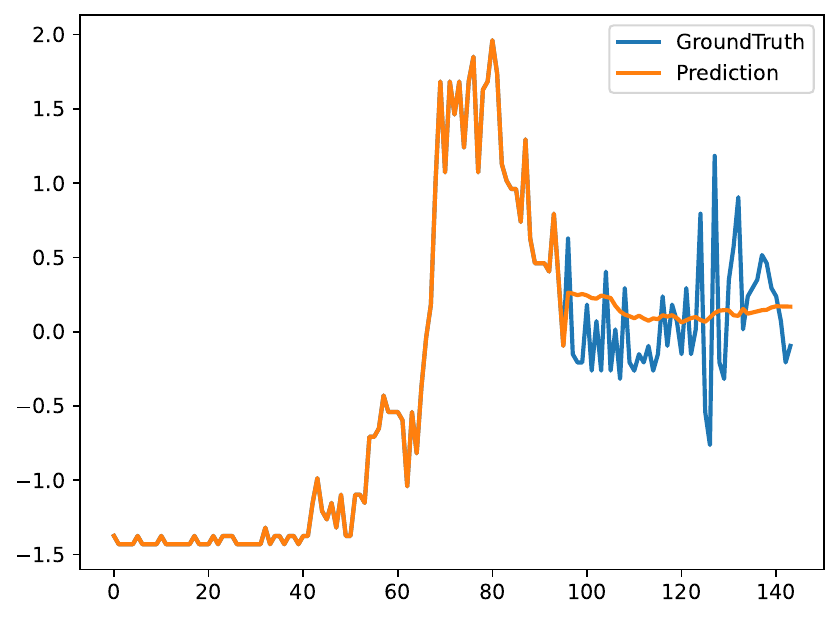}
    \caption{PatchTST} 
  \end{subfigure}
     \hspace{1em}
  \begin{subfigure}{0.3\columnwidth}
    \includegraphics[width=\textwidth]{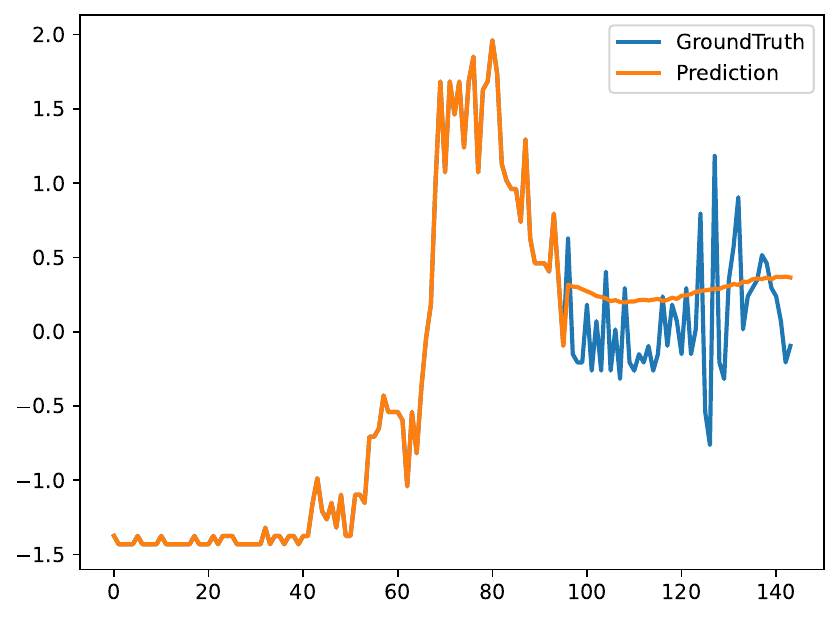}
    \caption{iTransformer} 
  \end{subfigure}
  
  \caption{Comparison on prediction graphs of PEMS08 dataset for prediction length 48}
  \label{fig-PEMS08-compare}
\end{figure}

We have provided comprehensive experiments in this area to assess the effectiveness of our suggested QCAAPatchTF approach in comparison to the most advanced time-series forecasting methods. To evaluate the effect of the proposed approach, we have also carried out an ablation investigation and hyperparameter sensitivity analysis. Every experiment is carried out on a single NVIDIA-GeForce RTX 3090 GPU using PyTorch and CUDA version 12.2. Table \ref{avglong} compares average multivariate long-term forecasting results across various datasets and multiple benchmark models. The results, based on average MSE and average MAE, highlight the performance of QCAAPatchTF, with achievements shown in red and blue colours, respectively. Figure \ref{predtestetth2} shows a sample long-term prediction comparison for our QCAAPatchTF approach and other benchmark models on the ETTh2 dataset. Table \ref{avgtabshort} compares the performance of five models (Crossformer, iTransformer, PatchTST, EDformer, and QCAAPatchTF) in multivariate short-term forecasting across four datasets (PEMS03, PEMS04, PEMS07, and PEMS08) using MSE and MAE metrics. QCAAPatchTF achieves the lowest MSE (highlighted in red) for the PEMS03 and PEMS07 datasets while also securing the best MAE values (highlighted in blue), demonstrating its competitive forecasting accuracy for the others. The sample short-term forecasts for the PEMS08 dataset are depicted in Figure \ref{fig-PEMS08-compare}. Additional results from the M4 dataset in Table \ref{table-resultsM4} further validate the competitiveness of QCAAPatchTF with other benchmark models. Furthermore, QCAAPatchTF's lightweight design and quantum parallel superposition technique ensure it delivers comparable results in less time. Table \ref{executionlong} and Table \ref{executionshort} present the average execution time (in seconds) for long-term and short-term forecasting tasks across various benchmark models, respectively. The results indicate that QCAAPatchTF is the second fastest model for these tasks, benefiting from the inherent parallelism of the proposed quantum-classical attention module. It is important to note that this comparison is influenced by the architecture of the underlying execution environment.

\begin{algorithm}[hbt!]
\scriptsize
\caption{Quantum Classical Self-Attention (QCSA)}
\begin{algorithmic}[1]
\Require Queries \( Q \in \mathbb{R}^{B \times L \times H \times E} \), Keys \( K \in \mathbb{R}^{B \times S \times H \times D} \), Values \( V \in \mathbb{R}^{B \times S \times H \times D} \), Optional Attention Mask \( M \)
\Ensure Attention-weighted output \( V_{\text{out}} \), optionally attention scores \( A \)

\State \textbf{Initialize} QuantumAttention module:
\State Set number of qubits \( n_q \) and entanglement factor \( \lambda \)

\State \textbf{Define Variational Quantum Circuit:}
\For{\( i = 1 \) to \( n_q \)}
    \State Apply parameterized rotation gate: \( R_Y(\theta_i) \)
\EndFor
\For{\( i = 1 \) to \( n_q - 1 \)}
    \State Apply entanglement via CNOT gate: \( CNOT(i, i+1) \)
\EndFor
\State Compute quantum expectation value: 
\[
S_{\text{quantum}} = \langle \psi(\theta) | Z | \psi(\theta) \rangle
\]
\State \textbf{Compute Attention Scores:}
\State Compute classical superposition-based scores:
\[
S_{\text{sup}} = Q K^T
\]
\State Compute quantum-based scores:
\[
S_{\text{quantum}} = \text{QuantumCircuit}(S_{\text{sup}})
\]
\State Compute entanglement-based scores:
\[
S_{\text{ent}} = V K^T
\]
\State Combine quantum and entanglement scores:
\[
S = S_{\text{quantum}} + \lambda S_{\text{ent}}
\]
\State \textbf{Apply Attention Mask (if enabled):}
\If{masking is enabled}
    \State Set \( S_{ij} = -\infty \) wherever \( M_{ij} = 1 \)
\EndIf
\State \textbf{Normalize Scores Using Softmax:}
\[
A = \text{softmax}(\alpha S), \quad \text{where } \alpha = \frac{1}{\sqrt{E}}
\]
\State \textbf{Compute Weighted Sum of Values:}
\[
V_{\text{out}} = A V
\]
\State \textbf{Return Output:}
\If{output attention is enabled}
    \Return \( V_{\text{out}}, A \)
\Else
    \Return \( V_{\text{out}} \)
\EndIf

\end{algorithmic}
\label{algo1}
\end{algorithm}

\begin{algorithm}[hbt!]
\scriptsize
\tiny
\caption{Quantum Classical Advanced Patch-based Transformer (QCAAPatchTF)}
\begin{algorithmic}
\State \textbf{Input:} Time Series Data $X_{\text{enc}}$, Time Marks $M_{\text{enc}}$, Decoding Data $X_{\text{dec}}$, Decoding Time Marks $M_{\text{dec}}$, Mask (optional)
\State Model Hyperparameters: Task Name, Sequence Length $L$, Prediction Length $P$, Dropout Rate $p$, Number of Heads $H$, etc.
\State \textbf{Output:} Task-Specific Prediction $Y$

\vspace{2mm}
\State \textbf{Step 1: Define Model Components}
\State Define Transpose Layer: 
\[
\text{Transpose}(\text{dims}, \text{contiguous}) = 
\begin{cases}
\text{Transpose}(\text{dims}), & \text{if contiguous is False} \\
\text{Transpose}(\text{dims}).\text{contiguous}(), & \text{otherwise}
\end{cases}
\]
\State Define FlattenHead Layer: 
\[
\text{FlattenHead}(n_{\text{vars}}, nf, \text{target\_window}, p) = \text{Flatten} \to \text{Linear Layer} \to \text{Dropout}
\]
\State Compute Optimized Patch Length and Stride. padding=stride.
\State Define Patch Embedding:
\text{PatchEmbedding}(\text{d\_model}, \text{optimized\_patch\_len}, \text{optimized\_stride}, \text{padding}, p)
\State Define Encoder with Attention Mechanism:
\[
\text{Encoder} = 
\begin{cases}
\text{QuantumAttention}(Q, K, V) = \frac{QK^T}{\sqrt{d}}V, & \text{if Quantum Attention is enabled} \\
\text{FullAttention}(Q, K, V) = \text{softmax}\left(\frac{QK^T}{\sqrt{d}}\right) V, & \text{otherwise}
\end{cases}
\]
\State \textbf{Step 2: Forecasting Task}
\If{Task = "forecasting"}
    \State Normalize $X_{\text{enc}}$:
    \[
    X_{\text{enc}}' = \frac{X_{\text{enc}} - \mu(X_{\text{enc}})}{\sqrt{\text{Var}(X_{\text{enc}}) + \epsilon}}
    \]   
    \State Apply Patch Embedding: 
    $
    Z_{\text{patch}}, n_{\text{vars}} = \text{PatchEmbedding}(X_{\text{enc}})
    $
    \State Apply Encoder: 
    $
    Z_{\text{enc}}, \text{attn} = \text{Encoder}(Z_{\text{patch}})
    $
    \State Reshape for Decoding:
    $
    Z_{\text{enc}} = \text{Reshape}(Z_{\text{enc}})
    $
    \State Apply Prediction Head:
    $
    Y_{\text{forecast}} = W_{\text{proj}} Z_{\text{enc}}
    $
\EndIf
\State \textbf{Step 3: Anomaly Detection}
\If{Task = "anomaly\_detection"}
    \State Normalize and Embed Data.
    \State Apply Encoder and Reshape.
    \State Compute Anomaly Score:
   $
    Y_{\text{anomaly}} = W_{\text{proj}} Z_{\text{enc}}
    $
\EndIf
\State \textbf{Step 4: Classification Task}
\If{Task = "classification"}
    \State Normalize and Embed Data.
    \State Apply Encoder and Reshape.
    \State Apply Activation and Dropout:
    $
    Z_{\text{act}} = \text{GELU}(Z_{\text{enc}}), \quad Z_{\text{drop}} = \text{Dropout}(Z_{\text{act}})
    $
    \State Flatten and Apply Projection:
    $
    Y_{\text{class}} = W_{\text{proj}} (\text{Flatten}(Z_{\text{drop}}))
    $
\EndIf
\State \textbf{Step 5: Output}
\If{Task = "forecasting"}
    \State Return forecast prediction: $Y_{\text{forecast}}$
\ElsIf{Task = "anomaly\_detection"}
    \State Return anomaly prediction: $Y_{\text{anomaly}}$
\ElsIf{Task = "classification"}
    \State Return classification prediction: $Y_{\text{class}}$
\EndIf

\end{algorithmic}
\label{algo2}
\end{algorithm}

\begin{table*}[hbt!]
\centering
\scriptsize
\tiny
\captionsetup{justification=centering}
\caption{Dataset descriptions}
\begin{tabular}{|c|c|c|c|c|c|}
\hline
Forecasting Type & Dataset      & Dim & Size                & Frequency & Information    \\ \hline
Long-term   & ETTh1, ETTh2 & 7   & (8545,2881,2881)    & Hourly    & Electricity    \\
            & ETTm1, ETTm2 & 7   & (34465,11521,11521) & 15 min    & Electricity    \\
            & Weather      & 21  & (36792,5271,10540)  & 10 min    & Weather        \\
            & Electricity  & 321 & (18317,2633,5261)   & Hourly    & Electricity    \\
            & Traffic      & 862 & (12185,1757,3509)   & Hourly    & Transportation \\
            & Exchange     & 8   & (5120,665,1422)     & Daily     & Economy        \\ \hline
Short-term \cite{pems_dataset}  & PEMS03       & 358 & (15617,5135,5135)   & 5 min     & Transportation \\
            & PEMS04       & 307 & (10172,3375,3375)   & 5 min     & Transportation \\
            & PEMS07       & 883 & (16911,5622,5622)   & 5 min     & Transportation \\
            & PEMS08       & 170 & (10690,3548,3548)   & 5 min     & Transportation \\ \hline
Classification (UEA)   
             & EthanolConcentration & 3   & (261, 0, 263)    & -   & Alcohol Industry
    \\
            & Handwriting   & 3   & (150, 0, 850)   & -    & Handwriting   \\
            & Heartbeat      & 61  & (204, 0, 205)   &-    &Heartbeat rate       \\
            & FaceDetection  & 144 & (5890, 0, 3524)    & 250 Hz    & Face    \\
            &  JapaneseVowels  & 12 & (270, 0, 370)   & -    & Voice \\
            & UWaveGestureLibrary   & 3   & (120, 0, 320)     & -    & Gesture   \\
            & SpokenArabicDigits   & 13   & (6599, 0, 2199)      & 11025 Hz    & Voice
            \\ \hline
Anomaly Detection  
            & SMD & 38   & (566724, 141681, 708420)    & -   & Server Machine
    \\
            & MSL   & 55   & (44653, 11664, 73729)   & -    & Spacecraft   \\
            & SMAP  & 25  & (108146, 27037, 427617)   &-    & Spacecraft        \\
            & SWaT  & 51  & (396000, 99000, 449919)     & -    & Infrastructure   \\
            & PSM  & 25  & (105984, 26497, 87841)    & -    & Server Machine 
            \\ \hline
\end{tabular}
\label{table-dataset}
\end{table*}

\begin{table}[hbt!]
\centering
\scriptsize
\tiny
\captionsetup{justification=centering}
\caption{Details of M4 data series.}
\begin{tabular}{|c|c|c|c|c|c|c|c|}
\hline
Time intervals & Micro           & Industry        & Macro           & Finance         & Demographic    & Other          & Total            \\ \hline
Yearly                                        & 6,538           & 3,716           & 3,903           & 6,519           & 1,088          & 1,236          & \textbf{23,000}  \\ 
Quarterly                                     & 6,020           & 4,637           & 5,315           & 5,305           & 1,858          & 865            & \textbf{24,000}  \\ 
Monthly                                       & 10,975          & 10,017          & 10,016          & 10,987          & 5,728          & 277            & \textbf{48,000}  \\ 
Weekly                                        & 112             & 6               & 41              & 164             & 24             & 12             & \textbf{359}     \\
Daily                                         & 1,476           & 422             & 127             & 1,559           & 10             & 633            & \textbf{4,227}   \\ 
Hourly                                        & 0               & 0               & 0               & 0               & 0              & 414            & \textbf{414}     \\ 
Total                                         & \textbf{25,121} & \textbf{18,798} & \textbf{19,402} & \textbf{24,534} & \textbf{8,708} & \textbf{3,437} & \textbf{100,000} \\ \hline
\end{tabular}
\label{M4dataset}
\end{table}

\begin{table}[hbt!]
\scriptsize
\tiny
\caption{Hyperparameters of QCAAPatchTF Approach for forecasting, classification and anomaly detection tasks}
\begin{center}
\captionsetup{justification=centering}
\begin{tabular}{|l|c|c|c|c|}
\hline
Parameter        & Long-term  & Short-term  &Classification   &Anomaly detection\\ \hline
d\_model                  & 512                 & 128    & 128      & 128       \\ \hline
channel\_independence     & 0 (except Exchange) & 0       & 0     & 0            \\ \hline
Number of scales ($k$)      & 4                   & 4      & 3     &  3/5            \\ \hline
Batch size                & 32                  & 16/32       & 16     & 128           \\ \hline
Learning rate             & 0.001              & 0.001/0.003       & 0.001     &  0.0001       \\ \hline
Patience (early stopping) & 3                   & 3          & 10     & 3         \\ \hline
Number of epochs          & 10                  & 10       & 50     & 10          \\ \hline
\end{tabular}
\label{table-impl}
\end{center}
\end{table}

\begin{table*}[hbt!]
\scriptsize
\tiny
\caption{Comparison of average error coefficients on multivariate long-term forecasting (prediction lengths - 96, 192, 336, 720). The red colour values provide the best average MSE and the blue colour values provide the best average MAE values.}
\begin{center}
\begin{tabular}{|c|cc|cc|cc|cc|cc|cc|cc|cc|cc|cc|}
\hline
Models      & \multicolumn{2}{c|}{Autoformer}                                                  & \multicolumn{2}{c|}{Informer}                                                    & \multicolumn{2}{c|}{NS-Transformer}                                              & \multicolumn{2}{c|}{Reformer}                                                    & \multicolumn{2}{c|}{Crossformer}                                                 & \multicolumn{2}{c|}{ETSFormer}                                                   & \multicolumn{2}{l|}{iTransformer}                                                & \multicolumn{2}{c|}{PatchTST}                                                    & \multicolumn{2}{c|}{FEDformer}                                                   & \multicolumn{2}{c|}{\textbf{QCAAPatchTF (Ours)}}                                 \\ \hline
Database    & MSE                                               & MAE                          & MSE                                               & MAE                          & MSE                                               & MAE                          & MSE                                               & MAE                          & \multicolumn{1}{c|}{MSE}                          & MAE                          & \multicolumn{1}{c|}{MSE}                          & MAE                          & \multicolumn{1}{c|}{MSE}                          & MAE                          & \multicolumn{1}{c|}{MSE}                          & MAE                          & \multicolumn{1}{c|}{MSE}                          & MAE                          & \multicolumn{1}{c|}{MSE}                          & MAE                          \\ \hline
ETTh1       & \multicolumn{1}{c|}{{\color[HTML]{000000} 0.504}} & {\color[HTML]{000000} 0.492} & \multicolumn{1}{c|}{{\color[HTML]{000000} 1.058}} & {\color[HTML]{000000} 0.808} & \multicolumn{1}{c|}{{\color[HTML]{000000} 0.609}} & {\color[HTML]{000000} 0.541} & \multicolumn{1}{c|}{{\color[HTML]{000000} 1.019}} & {\color[HTML]{000000} 0.763} & \multicolumn{1}{c|}{{\color[HTML]{000000} 0.557}} & {\color[HTML]{000000} 0.537} & \multicolumn{1}{c|}{{\color[HTML]{000000} 0.610}} & {\color[HTML]{000000} 0.582} & \multicolumn{1}{c|}{{\color[HTML]{000000} 0.450}} & {\color[HTML]{000000} 0.457} & \multicolumn{1}{c|}{{\color[HTML]{000000} 0.457}} & {\color[HTML]{3531FF} 0.453} & \multicolumn{1}{c|}{{\color[HTML]{FE0000} 0.439}} & {\color[HTML]{000000} 0.458} & \multicolumn{1}{c|}{{\color[HTML]{000000} 0.458}} & {\color[HTML]{000000} 0.454} \\ 
ETTh2       & \multicolumn{1}{c|}{{\color[HTML]{000000} 0.447}} & {\color[HTML]{000000} 0.463} & \multicolumn{1}{c|}{{\color[HTML]{000000} 4.665}} & {\color[HTML]{000000} 1.771} & \multicolumn{1}{c|}{{\color[HTML]{000000} 0.567}} & {\color[HTML]{000000} 0.509} & \multicolumn{1}{c|}{{\color[HTML]{000000} 2.604}} & {\color[HTML]{000000} 1.257} & \multicolumn{1}{c|}{{\color[HTML]{000000} 2.768}} & {\color[HTML]{000000} 1.324} & \multicolumn{1}{c|}{{\color[HTML]{000000} 0.441}} & {\color[HTML]{000000} 0.455} & \multicolumn{1}{c|}{{\color[HTML]{000000} 0.394}} & {\color[HTML]{000000} 0.413} & \multicolumn{1}{c|}{{\color[HTML]{000000} 0.393}} & {\color[HTML]{000000} 0.415} & \multicolumn{1}{c|}{{\color[HTML]{000000} 0.442}} & {\color[HTML]{000000} 0.454} & \multicolumn{1}{c|}{{\color[HTML]{FE0000} 0.380}} & {\color[HTML]{3531FF} 0.407} \\ 
ETTm1       & \multicolumn{1}{c|}{{\color[HTML]{000000} 0.571}} & {\color[HTML]{000000} 0.513} & \multicolumn{1}{c|}{{\color[HTML]{000000} 0.890}} & {\color[HTML]{000000} 0.701} & \multicolumn{1}{c|}{{\color[HTML]{000000} 0.521}} & {\color[HTML]{000000} 0.472} & \multicolumn{1}{c|}{{\color[HTML]{000000} 1.021}} & {\color[HTML]{000000} 0.731} & \multicolumn{1}{c|}{{\color[HTML]{000000} 0.591}} & {\color[HTML]{000000} 0.567} & \multicolumn{1}{c|}{\color{red}{0.304}} & {\color{blue}{0.359}} & \multicolumn{1}{c|}{{\color[HTML]{000000} 0.406}} & {\color[HTML]{000000} 0.411} & \multicolumn{1}{c|}{{\color[HTML]{000000} 0.365}} & {\color[HTML]{000000} 0.391} & \multicolumn{1}{c|}{{\color[HTML]{000000} 0.449}} & {\color[HTML]{000000} 0.457} & \multicolumn{1}{c|}{{\color[HTML]{000000} 0.388}} & {\color[HTML]{000000} 0.404} \\ 
ETTm2       & \multicolumn{1}{c|}{{\color[HTML]{000000} 0.338}} & {\color[HTML]{000000} 0.368} & \multicolumn{1}{c|}{{\color[HTML]{000000} 1.716}} & {\color[HTML]{000000} 0.903} & \multicolumn{1}{c|}{{\color[HTML]{000000} 0.642}} & {\color[HTML]{000000} 0.500} & \multicolumn{1}{c|}{{\color[HTML]{000000} 2.010}} & {\color[HTML]{000000} 1.034} & \multicolumn{1}{c|}{{\color[HTML]{000000} 1.296}} & {\color[HTML]{000000} 0.719} & \multicolumn{1}{c|}{{\color[HTML]{000000} 0.292}} & {\color[HTML]{000000} 0.349} & \multicolumn{1}{c|}{{\color[HTML]{000000} 0.290}} & {\color[HTML]{000000} 0.332} & \multicolumn{1}{c|}{{\color[HTML]{000000} 0.292}} & {\color[HTML]{3531FF} 0.334} & \multicolumn{1}{c|}{{\color[HTML]{000000} 0.307}} & {\color[HTML]{000000} 0.351} & \multicolumn{1}{c|}{{\color[HTML]{FE0000} 0.289}} & {\color[HTML]{3531FF} 0.334} \\
Weather     & \multicolumn{1}{c|}{{\color[HTML]{000000} 0.379}} & {\color[HTML]{000000} 0.407} & \multicolumn{1}{c|}{{\color[HTML]{000000} 0.627}} & {\color[HTML]{000000} 0.547} & \multicolumn{1}{c|}{{\color[HTML]{000000} 0.280}} & {\color[HTML]{000000} 0.314} & \multicolumn{1}{c|}{{\color[HTML]{000000} 0.535}} & {\color[HTML]{000000} 0.521} & \multicolumn{1}{c|}{{\color[HTML]{000000} 0.265}} & {\color[HTML]{000000} 0.327} & \multicolumn{1}{c|}{{\color[HTML]{000000} 0.263}} & {\color[HTML]{000000} 0.319} & \multicolumn{1}{c|}{{\color[HTML]{000000} 0.255}} & {\color[HTML]{000000} 0.281} & \multicolumn{1}{c|}{{\color[HTML]{000000} 0.257}} & {\color[HTML]{000000} 0.279} & \multicolumn{1}{c|}{{\color[HTML]{000000} 0.312}} & {\color[HTML]{000000} 0.364} & \multicolumn{1}{c|}{{\color[HTML]{FE0000} 0.254}} & {\color[HTML]{3531FF} 0.277} \\ 
Electricity & \multicolumn{1}{c|}{{\color[HTML]{000000} 0.255}} & {\color[HTML]{000000} 0.355} & \multicolumn{1}{c|}{{\color[HTML]{000000} 0.362}} & {\color[HTML]{000000} 0.439} & \multicolumn{1}{c|}{{\color[HTML]{000000} 0.199}} & {\color[HTML]{000000} 0.294} & \multicolumn{1}{c|}{{\color[HTML]{000000} 0.331}} & {\color[HTML]{000000} 0.410} & \multicolumn{1}{c|}{{\color[HTML]{000000} 0.278}} & {\color[HTML]{000000} 0.340} & \multicolumn{1}{c|}{{\color[HTML]{000000} 0.207}} & {\color[HTML]{000000} 0.323} & \multicolumn{1}{c|}{{\color[HTML]{FE0000} 0.181}} & {\color[HTML]{3531FF} 0.270} & \multicolumn{1}{c|}{{\color[HTML]{000000} 0.212}} & {\color[HTML]{000000} 0.309} & \multicolumn{1}{c|}{{\color[HTML]{000000} 0.295}} & {\color[HTML]{000000} 0.385} & \multicolumn{1}{c|}{{\color[HTML]{000000} 0.210}} & {\color[HTML]{000000} 0.301} \\ 
Traffic     & \multicolumn{1}{c|}{{\color[HTML]{000000} 0.661}} & {\color[HTML]{000000} 0.408} & \multicolumn{1}{c|}{{\color[HTML]{000000} 0.862}} & {\color[HTML]{000000} 0.487} & \multicolumn{1}{c|}{{\color[HTML]{000000} 0.648}} & {\color[HTML]{000000} 0.356} & \multicolumn{1}{c|}{{\color[HTML]{000000} 0.709}} & {\color[HTML]{000000} 0.391} & \multicolumn{1}{c|}{{\color[HTML]{000000} 0.563}} & {\color[HTML]{000000} 0.304} & \multicolumn{1}{c|}{{\color[HTML]{000000} 0.620}} & {\color[HTML]{000000} 0.395} & \multicolumn{1}{c|}{{\color[HTML]{FE0000} 0.444}} & {\color[HTML]{3531FF} 0.301} & \multicolumn{1}{c|}{{\color[HTML]{000000} 0.532}} & {\color[HTML]{000000} 0.342} & \multicolumn{1}{c|}{{\color[HTML]{000000} 0.615}} & {\color[HTML]{000000} 0.383} & \multicolumn{1}{c|}{{\color[HTML]{000000} 0.508}} & {\color[HTML]{000000} 0.333} \\ 
Exchange    & \multicolumn{1}{c|}{{\color[HTML]{000000} 0.628}} & {\color[HTML]{000000} 0.554} & \multicolumn{1}{c|}{{\color[HTML]{000000} 1.621}} & {\color[HTML]{000000} 1.005} & \multicolumn{1}{c|}{{\color[HTML]{000000} 0.505}} & {\color[HTML]{000000} 0.476} & \multicolumn{1}{c|}{{\color[HTML]{000000} 1.536}} & {\color[HTML]{000000} 1.013} & \multicolumn{1}{c|}{{\color[HTML]{000000} 0.755}} & {\color[HTML]{000000} 0.645} & \multicolumn{1}{c|}{{\color[HTML]{000000} 0.410}} & {\color[HTML]{000000} 0.427} & \multicolumn{1}{c|}{{\color[HTML]{000000} 0.387}} & {\color[HTML]{000000} 0.418} & \multicolumn{1}{c|}{{\color[HTML]{000000} 0.393}} & {\color[HTML]{000000} 0.418} & \multicolumn{1}{c|}{{\color[HTML]{000000} 0.520}} & {\color[HTML]{000000} 0.502} & \multicolumn{1}{c|}{{\color[HTML]{FE0000} 0.380}} & {\color[HTML]{3531FF} 0.410} \\ \hline
\end{tabular}
\label{avglong}
\end{center}
\end{table*}
  
\begin{table*}[hbt!]
\centering
\scriptsize
\tiny
\captionsetup{justification=centering}
\caption{Comparison of error coefficients on multivariate short-term forecasting (prediction length 48 and lookback 96).  The red colour values provide the best MSE and the blue colour values provide the best MAE values.}
\begin{tabular}{|c|cc|cc|cc|cc|cc|}
\hline
Models   & \multicolumn{2}{c|}{Crossformer}                            & \multicolumn{2}{c|}{iTransformer}                           & \multicolumn{2}{c|}{PatchTST}        & \multicolumn{2}{c|}{EDformer}                               & \multicolumn{2}{c|}{\textbf{QCAAPatchTF (Our)}}             \\ \hline
Database & MSE                          & MAE                          & MSE                          & MAE                          & MSE                          & MAE   & MSE                          & MAE                          & MSE                          & MAE                          \\ \hline
PEMS03   & 0.287                        & 0.393                        & 0.241                        & 0.343                        & {0.240} & 0.337 & 0.249                        & 0.345                        & {\color[HTML]{FE0000} 0.239} & {\color[HTML]{3531FF} 0.335} \\
PEMS04   & 0.241                        & 0.355                        & {\color[HTML]{FE0000} 0.218} & {\color[HTML]{3531FF} 0.319} & 0.313                        & 0.387 & 0.227                        & 0.344                        & 0.310                        & 0.386                        \\
PEMS07   & 0.295                        & 0.381                        & 0.274                        & 0.369                        & 0.298                        & 0.379 & {0.270} & {0.359}    & \color{red}{0.268}                        & \color{blue}{0.353}                        \\
PEMS08   & {\color[HTML]{FE0000} 0.210} & {\color[HTML]{3531FF} 0.265} & 0.237                        & 0.323                        & 0.268                        & 0.350 & 0.371                        & 0.446                        & 0.280                        & 0.359                        \\ \hline
\end{tabular}
\label{avgtabshort}
\end{table*}

\begin{table*}[hbt!]
\scriptsize
\tiny
\caption{Summary of short-term forecasting results on M4 dataset. Every prediction length may be found in [6, 48]. Red colour values highlight the best average results, and blue colour values indicate the second best.}
\begin{center}
\begin{tabular}{|l|c|c|c|c|c|c|c|c|c|}
\hline
\textbf{Metric}                  & \textbf{Category} & \textbf{EDformer}                      & \textbf{iTransformer} & \textbf{Reformer}                      & \textbf{NS-Transformer} & \textbf{Informer} & \textbf{Autoformer} & \textbf{Crossformer} & \textbf{QCAAPatchTF (Our)} \\ \hline
                                 & Yearly            & 14.259                                 & 14.409                & 14.548                                 & 15.833                  & 15.215            & 16.909              & 69.344               & 13.593         \\ 
                                 & Quarterly         & 11.407                                 & 10.777                & 11.922                                 & 12.366                  & 12.696            & 14.445              & 73.585               & 10.779          \\  
                                 & Monthly           & 15.558                                 & 16.650                & 14.649                                 & 14.607                  & 15.210            & 18.280              & 69.80                &  14.094         \\ 
                                 & Others            & 5.222                                  & 5.543                 & 6.694                                  & 7.005                   & 7.183             & 6.676               & 98.492               & 5.693           \\ \hline
\multirow{-5}{*}{\textbf{sMAPE}} & \textbf{Average}  & \textcolor{blue}{\textbf{13.796}} & \textbf{14.170}       & \textbf{14.192}                        & \textbf{14.201}         & \textbf{14.206}   & \textbf{16.464}     & \textbf{72.038}      & \textcolor{red}{\textbf{12.763}} \\ \hline
                                 & Yearly            & 17.558                                 & 19.191                & 17.789                                 & 20.485                  & 19.837            & 23.266              & 61.950               & 17.158          \\  
                                 & Quarterly         & 13.006                                 & 12.871                & 12.737                                 & 14.490                  & 14.969            & 16.882              & 66.971               & 12.808          \\ 
                                 & Monthly           & 18.318                                 & 20.144                & 15.830                                 & 16.988                  & 17.972            & 22.442              & 68.507               & 16.749          \\ 
                                 & Others            & 7.142                                  & 7.750                 & 10.456                                 & 10.459                  & 10.469            & 11.146              & 64.928               & 10.374         \\  \hline
\multirow{-5}{*}{\textbf{MAPE}}  & \textbf{Average}  & \textbf{16.409}                        & \textbf{17.560}       & {\color[HTML]{FE0000} \textbf{14.971}} & \textbf{16.689}         & \textbf{17.305}   & \textbf{20.732}     & \textbf{66.451}      & \textcolor{blue}{\textbf{15.578}} \\ \hline
                                 & Yearly            & 3.158                                  & 3.218                 & 3.232                                  & 3.532                   & 3.398             & 3.761               & 18.11                & 3.047          \\ 
                                 & Quarterly         & 1.426                                  & 1.284                 & 1.313                                  & 1.519                   & 1.561             & 1.854               & 13.313               & 1.278           \\
                                 & Monthly           & 1.189                                  & 1.392                 & 1.262                                  & 1.177                   & 1.217             & 1.572               & 11.168               & 1.127          \\ 
                                 & Others            & 4.568                                  & 3.998                 & 4.424                                  & 4.691                   & 4.937             & 4.833               & 79.686               & 3.694         \\ \hline
\multirow{-5}{*}{\textbf{MASE}}  & \textbf{Average}  & \textcolor{blue}{\textbf{1.868}}  & \textbf{1.916}        & \textbf{1.894}                         & \textbf{1.910}          & \textbf{1.987}    & \textbf{2.306}      & \textbf{16.705}      & \textcolor{red}{\textbf{1.733}}  \\ \hline
                                 & Yearly            & 0.834                                  & 0.846                 & 0.796                                  & 0.929                   & 0.893             & 0.991               & 4.40                 & 0.799          \\
                                 & Quarterly         & 1.038                                  & 0.957                 & 0.975                                  & 1.115                   & 1.145             & 1.332               & 8.195                & 0.955           \\
                                 & Monthly           & 1.098                                  & 1.232                 & 0.972                                  & 1.060                   & 1.099             & 1.373               & 7.670                &  1.018         \\
                                 & Others            & 1.375                                  & 1.214                 & 1.402                                  & 1.502                   & 1.534             & 1.465               & 22.930               & 1.181           \\ \hline
\multirow{-5}{*}{\textbf{OWA}}   & \textbf{Average}  & \textbf{0.997}                         & \textbf{1.023}        & {\color[HTML]{FE0000} \textbf{0.921}}  & \textbf{1.039}          & \textbf{1.043}    & \textbf{1.210}      & \textbf{7.024}       & \textcolor{blue}{\textbf{0.924}}  \\ \hline
\end{tabular}
\label{table-resultsM4}
\end{center}
\end{table*}

\begin{table*}[hbt!]
\scriptsize
\tiny
\caption{Comparison of the execution time (seconds) of multivariate long-term forecasting results. The red colour values represent the lowest average execution time and the blue values represent the second lowest.}
\begin{center}
\begin{tabular}{|c|c|c|c|c|c|c|c|c|}
\hline
\multicolumn{1}{|l|}{}                & Datasets & Autoformer & Informer & Reformer & NS-Trans & iTransformer &PatchTST    & \textbf{QCAAPatchTF(Our)} \\ \hline
\multirow{4}{*}{Long-term Time (Sec)} & ETTh1    & 2.715      & 1.591    & 1.928      & 1.220    & \textcolor{red}{0.851}   &1.102     & \textcolor{blue}{0.985}           \\  
                                      & ETTh2    & 4.242      & 1.551    & 1.936      & 1.315    & \textcolor{red}{0.769}    &1.335    & \textcolor{blue}{1.126}           \\
                                      & Weather  & 12.462     & 5.423    & 6.711      & 5.831    & \textcolor{red}{3.374}    & 4.911   & \textcolor{blue}{4.321}           \\ 
                                      & Exchange & 1.826      & 0.928    & 1.114     & 0.991    & \textcolor{red}{0.539}    & 0.791   & \textcolor{blue}{0.639}           \\ \hline
\end{tabular}
\label{executionlong}
\end{center}
\end{table*}

\begin{table}[hbt!]
\scriptsize
\tiny
\caption{Comparison of the execution time (seconds) of multivariate short-term forecasting results. The red colour values represent the lowest average execution time and the blue colour values represent the second lowest.}
\begin{center}
\begin{tabular}{|c|c|c|c|c|}
\hline
Datasets & Crossformer & PatchTST  & iTransformer & \textbf{QCAAPatchTF(Our)} \\ \hline
PEMS03   & 14.345   & 9.791      & \textcolor{red}{5.887}        & \textcolor{blue}{8.617}                          \\
PEMS04   & 7.996   & 6.907        & \textcolor{red}{3.803}        & \textcolor{blue}{5.982}                          \\
PEMS07   & 38.76   & 30.597      & \textcolor{red}{22.755}       & \textcolor{blue}{28.230}                           \\ 
PEMS08   & 5.382    & 3.849       & \textcolor{red}{1.388}        & \textcolor{blue}{3.832}                     \\ \hline
\end{tabular}
\label{executionshort}
\end{center}
\end{table}

\vspace{-10pt}
\vspace{-10pt}

\subsection{Classification}
This study employs sequence-level classification. Seven multivariate datasets from the UEA Time Series Classification Archive \cite{bagnall2018uea} are selected, spanning applications such as gesture, face, and audio recognition, as well as heartbeat-based medical diagnosis. To ensure consistency across varying sequence lengths, the datasets are preprocessed following \cite{zerveas2021transformer}. Table \ref{classfull} provides a detailed comparison of classification accuracy (\%) across multiple datasets and models, with the best accuracy for each dataset highlighted in red. Notable observations include Crossformer achieving the highest accuracy (33.0\%) for EthanolConcentration, Transformer excelling in Handwriting (37.5\%), and iTransformer leading in Heartbeat (75.3\%). QCAAPatchTF demonstrates superior performance in FaceDetection (68.7\%) and UWaveGestureLibrary (86.7\%), while Crossformer achieves the highest accuracy for JapaneseVowels (97.6\%). Reformer outperforms others on SpokenArabicDigits (98.7\%). These results highlight the varying performance of models based on dataset characteristics, with QCAAPatchTF showing strong accuracy in specific cases due to its enhanced expressiveness, parallel computation capabilities, and adaptive variational quantum principles.

\subsection{Anomaly Detection}
Detecting anomalies in monitoring data is crucial for effective industrial maintenance \cite{xu2021anomaly}. However, anomalies are often hidden within large-scale datasets, making manual labeling a significant challenge. To address this, we have focused on unsupervised time series anomaly detection, enabling the identification of abnormal time points without the need for labeled data. We have evaluated models on five widely used anomaly detection benchmarks: SMD, MSL, SMAP, SWaT, and PSM. Table \ref{anofull} presents precision (P), recall (R), and F1-score (F1) across six anomaly detection datasets, where higher values indicate better performance. The proposed QCAAPatchTF model achieves the highest F1-scores in some cases, particularly on SMD (81.5\%) and PSM (96.3\%), demonstrating its competitive ability to balance precision and recall. While SWaT, MSL, and SMAP yield competitive results across multiple models, QCAAPatchTF remains highly effective, reinforcing its robustness in anomaly detection.

\begin{table*}[hbt!]
\scriptsize
\tiny
\caption{Full summary of classification results in terms of classification accuracy (\%). The red colour values denote the best accuracy.}
\begin{center}
\begin{tabular}{|c|ccccccccc|}
\hline
                     & \multicolumn{9}{c|}{Models}                                                                                                                                                                                                                                                                                                                                                \\ \hline
Dataset              & \multicolumn{1}{c|}{Autoformer} & \multicolumn{1}{c|}{Informer} & \multicolumn{1}{c|}{Reformer}                    & \multicolumn{1}{c|}{iTransformer}                & \multicolumn{1}{c|}{FEDformer} & \multicolumn{1}{c|}{Crossformer}                 & \multicolumn{1}{c|}{PatchTST} & \multicolumn{1}{c|}{Transformer}                 & \textbf{QCAAPatchTF (Our)}  \\ \hline
EthanolConcentration & \multicolumn{1}{c|}{28.9}       & \multicolumn{1}{c|}{28.1}     & \multicolumn{1}{c|}{28.8}                        & \multicolumn{1}{c|}{27.7}                        & \multicolumn{1}{c|}{28.5}      & \multicolumn{1}{c|}{{\color[HTML]{FE0000} 33.0}} & \multicolumn{1}{c|}{28.1}     & \multicolumn{1}{c|}{28.1}                        & 25.8                        \\
Handwriting          & \multicolumn{1}{c|}{18.6}       & \multicolumn{1}{c|}{32.0}     & \multicolumn{1}{c|}{31.2}                        & \multicolumn{1}{c|}{26.1}                        & \multicolumn{1}{c|}{23.5}      & \multicolumn{1}{c|}{29.1}                        & \multicolumn{1}{c|}{26.5}     & \multicolumn{1}{c|}{{\color[HTML]{FE0000} 37.5}} & 25.7                        \\
Heartbeat            & \multicolumn{1}{c|}{72.2}       & \multicolumn{1}{c|}{75.0}     & \multicolumn{1}{c|}{75.1}                        & \multicolumn{1}{c|}{{\color[HTML]{FE0000} 75.3}} & \multicolumn{1}{c|}{75.1}      & \multicolumn{1}{c|}{75.0}                        & \multicolumn{1}{c|}{70.7}     & \multicolumn{1}{c|}{71.5}                        & 70.8                        \\
FaceDetection        & \multicolumn{1}{c|}{65.7}       & \multicolumn{1}{c|}{67.2}     & \multicolumn{1}{c|}{68.2}                        & \multicolumn{1}{c|}{66.2}                        & \multicolumn{1}{c|}{66.6}      & \multicolumn{1}{c|}{61.6}                        & \multicolumn{1}{c|}{67.3}     & \multicolumn{1}{c|}{67.8}                        & {\color[HTML]{FE0000} 68.7} \\
JapaneseVowels       & \multicolumn{1}{c|}{96.4}       & \multicolumn{1}{c|}{97.0}     & \multicolumn{1}{c|}{97.0}                        & \multicolumn{1}{c|}{97.0}                        & \multicolumn{1}{c|}{97.3}      & \multicolumn{1}{c|}{{\color[HTML]{FE0000} 97.5}} & \multicolumn{1}{c|}{95.9}     & \multicolumn{1}{c|}{97.0}                        & 95.4                        \\
UWaveGestureLibrary  & \multicolumn{1}{c|}{51.2}       & \multicolumn{1}{c|}{85.2}     & \multicolumn{1}{c|}{86.3} & \multicolumn{1}{c|}{85.7}                        & \multicolumn{1}{c|}{60.9}      & \multicolumn{1}{c|}{85.3}                        & \multicolumn{1}{c|}{86.2}     & \multicolumn{1}{c|}{86.5}                        & {\color[HTML]{FE0000}86.7}                        \\
SpokenArabicDigits   & \multicolumn{1}{c|}{97.9}       & \multicolumn{1}{c|}{98.6}     & \multicolumn{1}{c|}{{\color[HTML]{FE0000} 98.7}} & \multicolumn{1}{c|}{98.0}                        & \multicolumn{1}{c|}{98.4}      & \multicolumn{1}{c|}{96.7}                        & \multicolumn{1}{c|}{96.4}     & \multicolumn{1}{c|}{98.4}                        & 97.0                        \\ \hline
\end{tabular}
\label{classfull}
\end{center}
\end{table*}

\begin{table*}[hbt!]
\scriptsize
\tiny
\caption{Complete results of the anomaly detection task. P, R, and F1 denote precision (\%), recall (\%), and F1-score (\%), respectively with anomaly\_ratio 1. Better performance is indicated by higher P (blue), R (orange), and F1 (red) values.}
\begin{center}
\begin{tabular}{|c|ccc|ccc|ccc|ccc|ccc|ccc|ccc|}
\hline
Dataset & \multicolumn{3}{c|}{Informer}                                                                                                     & \multicolumn{3}{c|}{iTransformer}                                                                          & \multicolumn{3}{c|}{Crossformer}                             & \multicolumn{3}{c|}{PatchTST}                                & \multicolumn{3}{c|}{Transformer}                                                                           & \multicolumn{3}{c|}{FEDformer}                                                                                                    & \multicolumn{3}{c|}{\textbf{QCAAPatchTF (Our)}}                                                                                   \\ \hline
Metrics & \multicolumn{1}{c|}{P}                           & \multicolumn{1}{c|}{R}                           & F1                          & \multicolumn{1}{c|}{P}    & \multicolumn{1}{c|}{R}                           & F1                          & \multicolumn{1}{c|}{P}    & \multicolumn{1}{c|}{R}    & F1   & \multicolumn{1}{c|}{P}    & \multicolumn{1}{c|}{R}    & F1   & \multicolumn{1}{c|}{P}                           & \multicolumn{1}{c|}{R}    & F1                          & \multicolumn{1}{c|}{P}                           & \multicolumn{1}{c|}{R}                           & F1                          & \multicolumn{1}{c|}{P}                           & \multicolumn{1}{c|}{R}                           & F1                          \\ \hline
SMD     & \multicolumn{1}{c|}{72.8}                        & \multicolumn{1}{c|}{84.8}                        & 78.3                        & \multicolumn{1}{c|}{76.8} & \multicolumn{1}{c|}{77.8}                        & 81.2                        & \multicolumn{1}{c|}{72.1} & \multicolumn{1}{c|}{84.4} & 77.8 & \multicolumn{1}{c|}{76.5} & \multicolumn{1}{c|}{86.1} & 81.0 & \multicolumn{1}{c|}{72.7}                        & \multicolumn{1}{c|}{84.8} & 78.3                        & \multicolumn{1}{c|}{72.7}                        & \multicolumn{1}{c|}{81.5}                        & 76.9                        & \multicolumn{1}{c|}{{\color[HTML]{3531FF} 76.9}} & \multicolumn{1}{c|}{{\color[HTML]{F56B00} 86.8}} & {\color[HTML]{FE0000} 81.5} \\
MSL     & \multicolumn{1}{c|}{90.1}                        & \multicolumn{1}{c|}{73.6}                        & 81.0                        & \multicolumn{1}{c|}{86.2} & \multicolumn{1}{c|}{62.6}                        & 72.4                        & \multicolumn{1}{c|}{90.3} & \multicolumn{1}{c|}{72.8} & 80.6 & \multicolumn{1}{c|}{88.5} & \multicolumn{1}{c|}{71.3} & 79.0 & \multicolumn{1}{c|}{89.6}                        & \multicolumn{1}{c|}{73.6} & 80.9                        & \multicolumn{1}{c|}{{\color[HTML]{3531FF} 90.6}} & \multicolumn{1}{c|}{{\color[HTML]{F56B00} 75.2}} & {\color[HTML]{FE0000} 82.2} & \multicolumn{1}{c|}{88.6}                        & \multicolumn{1}{c|}{71.6}                        & 79.2                        \\
SMAP    & \multicolumn{1}{c|}{90.6}                        & \multicolumn{1}{c|}{{\color[HTML]{F56B00} 61.7}} & {\color[HTML]{FE0000} 73.4} & \multicolumn{1}{c|}{90.6} & \multicolumn{1}{c|}{53.0}                        & 66.9                        & \multicolumn{1}{c|}{89.6} & \multicolumn{1}{c|}{53.6} & 67.1 & \multicolumn{1}{c|}{89.9} & \multicolumn{1}{c|}{53.7} & 67.3 & \multicolumn{1}{c|}{{\color[HTML]{3531FF} 91.0}} & \multicolumn{1}{c|}{61.5} & {\color[HTML]{FE0000} 73.4} & \multicolumn{1}{c|}{90.1}                        & \multicolumn{1}{c|}{55.4}                        & 68.6                        & \multicolumn{1}{c|}{90.1}                        & \multicolumn{1}{c|}{55.6}                        & 68.8                        \\
SWaT    & \multicolumn{1}{c|}{{\color[HTML]{3531FF} 99.7}} & \multicolumn{1}{c|}{68.1}                        & 80.9                        & \multicolumn{1}{c|}{92.2} & \multicolumn{1}{c|}{{\color[HTML]{F56B00} 93.1}} & {\color[HTML]{FE0000} 92.7} & \multicolumn{1}{c|}{97.7} & \multicolumn{1}{c|}{84.4} & 90.6 & \multicolumn{1}{c|}{90.9} & \multicolumn{1}{c|}{79.7} & 84.9 & \multicolumn{1}{c|}{99.6}                        & \multicolumn{1}{c|}{68.9} & 81.5                        & \multicolumn{1}{c|}{99.0}                        & \multicolumn{1}{c|}{68.2}                        & 80.7                        & \multicolumn{1}{c|}{90.9}                        & \multicolumn{1}{c|}{79.7}                        & 85.0           (Our)             \\
PSM     & \multicolumn{1}{c|}{98.7}                        & \multicolumn{1}{c|}{83.1}                        & 90.2                        & \multicolumn{1}{c|}{98.1} & \multicolumn{1}{c|}{93.1}                        & 95.5                        & \multicolumn{1}{c|}{97.3} & \multicolumn{1}{c|}{87.8} & 92.3 & \multicolumn{1}{c|}{99.0} & \multicolumn{1}{c|}{93.5} & 96.2 & \multicolumn{1}{c|}{99.5}                        & \multicolumn{1}{c|}{83.2} & 90.6                        & \multicolumn{1}{c|}{{\color[HTML]{3531FF} 99.9}} & \multicolumn{1}{c|}{81.8}                        & 90.0                        & \multicolumn{1}{c|}{99.1}                        & \multicolumn{1}{c|}{{\color[HTML]{F56B00} 93.7}} & {\color[HTML]{FE0000} 96.3} \\ \hline
\end{tabular}
\label{anofull}
\end{center}
\end{table*}

\section{SUPERIORITY OF QCSA MECHANISM}
Integrating the 'Quantum Classical Self-Attention (QCSA)' mechanism into a time series transformer offers several distinct advantages over traditional attention mechanisms. The hybrid quantum-classical approach takes advantage of quantum principles such as superposition, entanglement, and variational eigensolvers, providing a more expressive and efficient method for modeling complex dependencies in time series data. By combining quantum-based attention scores with classical methods, QCSA can enhance the model's capability to capture long-range dependencies, mitigate noise, and model intricate relationships between tokens. The incorporation of quantum circuits introduces parallel processing capabilities, potentially accelerating inference and improving scalability for large time series datasets. Additionally, QCSA can capture dynamic and nonlinear interactions between different components of the time series, which are often challenging for classical attention mechanisms. QCSA improves generalization, flexibility, and convergence but faces challenges like computational overhead and specialized hardware requirements. The detailed proofs of these Lemmas are described in the supplementary document.

\begin{lemma}
The 'Quantum Classical Self-Attention' mechanism, combining quantum superposition and quantum entanglement with classical attention scores, ensures that the attention weights remain non-negative (\(A \geq 0\)) and retain probabilistic structure (\(\sum_i A_{i} = 1\)) after normalization via softmax. This mechanism is further stabilized through layer normalization and dropout, ensuring efficient training and preventing overfitting.
\end{lemma}

\begin{lemma}
Let \( Q \in \mathbb{R}^{B \times L \times H \times E} \), \( K \in \mathbb{R}^{B \times S \times H \times D} \), and \( V \in \mathbb{R}^{B \times S \times H \times D} \) be the queries, keys, and values, respectively, and let \( S_{\text{sup}} = QK^T \) represent the classical attention scores. The quantum-based attention scores \( S_{\text{quantum}} \) are computed through a variational quantum circuit, where the entanglement factor \( \lambda \) controls the trade-off between classical and quantum contributions. The final attention scores are:
\[
S = S_{\text{quantum}} + \lambda S_{\text{ent}}
\]
These scores are normalized via the softmax function:
\[
A = \text{softmax}(\alpha S), \quad \alpha = \frac{1}{\sqrt{E}}
\]
where \( A \) represents the attention weights. The attention-weighted output is computed as:
$V_{\text{out}} = A V$
and the model converges to a stable fixed point due to the iterative update rule, ensuring efficient learning.
\end{lemma}

\section{ABLATION STUDY}
Table \ref{table-ablation} presents the performance of various model configurations across four datasets—ETTh2, ETTm2, Weather, and Exchange—evaluated over four prediction lengths (96, 192, 336, 720). The influence of incorporating the quantum-classical (QCA) hybrid attention mechanism and the patch embedding operation into the forecasting model is investigated in this work. The three main configurations, quantum-classical attention with patch embedding (QCA+OptPatch), full attention (FA) with optimized patch embedding (FA+OptPatch), and full attention without optimized patch embedding (FA+WOptPatch) are compared in the table. The findings show that the suggested QCA method and OptPatch embedding work together to produce the most performance gains, underscoring their crucial function in raising the forecasting accuracy of the model.

\begin{table}[hbt!]
\centering
\scriptsize
\tiny
\caption{Ablation Study: Comparison of multivariate long-term forecasting average results.}
\begin{tabular}{|c|c|c|c|c|c|c|}
\hline
\textbf{} & \textbf{\begin{tabular}[c]{@{}c@{}}QCA+OptPatch\end{tabular}} & \textbf{\begin{tabular}[c]{@{}c@{}}FA+OptPatch\end{tabular}} & \textbf{\begin{tabular}[c]{@{}c@{}}FA+WOptPatch\end{tabular}} & \textbf{MSE}     & \textbf{MAE}     \\ \hline
   & \checkmark & $-$ & $-$ & \textcolor{red}{0.380} & \textcolor{blue}{0.409} \\
\textbf{ETTh2}    & $-$ & \checkmark & $-$ & 0.393 & 0.415 \\
    & $-$ & $-$ & \checkmark & 0.409 & 0.426 \\ \hline
   & \checkmark & $-$ & $-$ & \textcolor{red}{0.289} & \textcolor{blue}{0.334} \\
\textbf{ETTm2}    & $-$ & \checkmark & $-$ & 0.292 & 0.334 \\
   & $-$ & $-$ & \checkmark & 0.311 & 0.352 \\ \hline
    & \checkmark & $-$ & $-$ & \textcolor{red}{0.254}  & \textcolor{blue}{0.277} \\
\textbf{Weather}    & $-$ & \checkmark & $-$ & 0.257  & 0.279 \\
    & $-$ & $-$ & \checkmark & 0.281  & 0.298 \\ \hline
    & \checkmark & $-$ & $-$ & \textcolor{red}{0.380}  & \textcolor{blue}{0.410} \\
\textbf{Exchange}    & $-$ & \checkmark & $-$ & 0.393  & 0.418 \\
    & $-$ & $-$ & \checkmark & 0.399  & 0.429 \\ \hline
\end{tabular}
\label{table-ablation}
\end{table}

\section{HYPERPARAMETER SENSITIVITY}
In the sensitivity analysis for the classification task, the optimal value of the hyperparameter k is chosen to assess how it impacts the accuracy of the QCAAPatchTF model. The results in Table \ref{hypertabel1} show that the accuracy of the QCAAPatchTF approaches fluctuates with changes in k, demonstrating the models' sensitivity to this hyperparameter. This analysis offers important insights into the stability and robustness of the approach across different k values, which will inform future optimization and hyperparameter tuning for improved classification performance. Additionally, we have assessed the sensitivity of QCAAPatchTF’s performance to varying learning rates as a crucial hyperparameter. Table \ref{hypertabel2} presents the sensitivity analysis of the QCAAPatchTF model across three learning rates (0.001, 0.003, and 0.005) on PEMS datasets for short-term forecasting, with a prediction horizon of 48 and a look-back window of 96. For most datasets, a learning rate of 0.003 yields the best performance, achieving the lowest error metrics, such as MSE and MAE. These findings indicate that 0.003 provides the optimal balance for model training, outperforming both the lower (0.001) and higher (0.005) learning rates. Given its significant impact on model convergence and stability, selecting an appropriate learning rate remains a crucial factor in optimizing performance. Table \ref{hypertabel3} presents the sensitivity analysis of hyperparameters associated with the QCAAPatchTF model with respect to the anomaly ratio in anomaly detection tasks across three datasets: SMD, SMAP, and PSM. The anomaly ratio directly impacts model performance, as reflected in the F1-Score, which balances precision and recall. A lower anomaly ratio (e.g., anomaly\_ratio = 1) assumes anomalies are rare, enforcing stricter detection thresholds. This typically enhances precision at the expense of recall, resulting in higher F1-Scores for datasets like SMD (81.5\%) and SMAP (68.8\%). Conversely, increasing the anomaly ratio to 2 or 3 relaxes the thresholds, improving recall but slightly reducing precision, leading to a marginal decline in F1-Score for datasets such as SMAP and SMD. These results highlight the need to carefully adjust the anomaly ratio to balance precision and recall for effective anomaly detection across various datasets. The red colour values denoted in these three tables are the optimum values that have been used in this experiment.

\begin{table}[hbt!]
\scriptsize
\tiny
\caption{Hyperparameter (k) Sensitivity analysis against accuracy (\%) in classification task.}
\begin{center}
\begin{tabular}{|ccccc|}
\hline
\multicolumn{5}{|c|}{\begin{tabular}[c]{@{}c@{}}Accuracy(\%) of QCAAPatchTF\end{tabular}}                                                                                                                                               \\ \hline
\multicolumn{1}{|c|}{Dataset}                    & \multicolumn{1}{c|}{Handwriting}                 & \multicolumn{1}{c|}{JapaneseVowels}              & \multicolumn{1}{c|}{UWaveGestureLibrary}         & SpokenArabicDigits          \\ \hline
\multicolumn{1}{|c|}{k=1}                        & \multicolumn{1}{c|}{25.6}                        & \multicolumn{1}{c|}{95.3}                        & \multicolumn{1}{c|}{84.6}                        & 96.9                        \\
\multicolumn{1}{|c|}{k=2}                        & \multicolumn{1}{c|}{25.7}                        & \multicolumn{1}{c|}{95.4}                        & \multicolumn{1}{c|}{84.6}                        & 97.0                        \\
\multicolumn{1}{|c|}{{\color[HTML]{FE0000} k=3}} & \multicolumn{1}{c|}{{\color[HTML]{FE0000} 25.7}} & \multicolumn{1}{c|}{{\color[HTML]{FE0000} 95.4}} & \multicolumn{1}{c|}{{\color[HTML]{FE0000} 84.7}} & {\color[HTML]{FE0000} 97.0} \\
\multicolumn{1}{|c|}{k=4}                        & \multicolumn{1}{c|}{25.6}                        & \multicolumn{1}{c|}{95.3}                        & \multicolumn{1}{c|}{84.7}                        & 96.8                        \\ \hline
\end{tabular}
\label{hypertabel1}
\end{center}
\end{table}

\begin{table}[hbt!]
\scriptsize
\tiny
\caption{Hyperparameter sensitivity analysis with respect to the learning rate (LR), for short-term forecasting (prediction length 48 and look-back 96). The red colour value is the optimum value.}
\begin{center}
\begin{tabular}{|c|cc|cc|cc|}
\hline
\textbf{QCAAPatchTF} & \multicolumn{2}{c|}{LR=0.001} & \multicolumn{2}{c|}{{\color[HTML]{FE0000} LR=0.003}} & \multicolumn{2}{c|}{LR=0.005} \\ \hline
Database             & MSE           & MAE           & MSE                       & MAE                      & MSE           & MAE           \\ \hline
PEMS03               & 0.248         & 0.342         & 0.239                     & 0.335                    & 0.269         & 0.359         \\
PEMS04               & 0.323         & 0.397         & 0.310                     & 0.386                    & 0.338         & 0.405         \\
PEMS07               & 0.299         & 0.378         & 0.269                     & 0.353                    & 0.321         & 0.410         \\
PEMS08               & 0.300         & 0.378         & 0.280                     & 0.359                    & 0.301         & 0.379         \\ \hline
\end{tabular}
\end{center}
\label{hypertabel2}
\end{table}

\begin{table}[hbt!]
\scriptsize
\tiny
\caption{Hyperparameter sensitivity analysis with respect to the anomaly ratio for anomaly detection. The red colour (anomaly\_ratio) value is the optimum value.}
\begin{center}
\begin{tabular}{|c|cl|cl|cl|}
\hline
\textbf{QCAAPatchTF} & \multicolumn{2}{c|}{{\color[HTML]{FE0000} anomaly\_ratio=1}} & \multicolumn{2}{c|}{anomaly\_ratio=2} & \multicolumn{2}{c|}{anomaly\_ratio=3} \\ \hline
Database             & \multicolumn{2}{c|}{F1\_Score(\%)}                               & \multicolumn{2}{c|}{F1\_Score(\%)}        & \multicolumn{2}{c|}{F1\_Score(\%)}        \\ \hline
SMD                  & \multicolumn{2}{c|}{81.5}                                    & \multicolumn{2}{c|}{76.2}             & \multicolumn{2}{c|}{70.3}             \\ \hline
SMAP                 & \multicolumn{2}{c|}{68.8}                                    & \multicolumn{2}{c|}{67.1}             & \multicolumn{2}{c|}{65.3}             \\ \hline
PSM                  & \multicolumn{2}{c|}{96.3}                                    & \multicolumn{2}{c|}{97.0}             & \multicolumn{2}{c|}{96.8}             \\ \hline
\end{tabular}
\end{center}
\label{hypertabel3}
\end{table}

\section{CONCLUSION}
\label{conclusions}
This work introduces a quantum-classical hybrid attention-based advanced patch transformer (QCAAPatchTF) for multivariate time series analysis. QCAAPatchTF integrates a quantum-classical hybrid attention mechanism within an optimized patch-based transformer framework, delivering consistent performance enhancements across benchmark architectures. Its versatility makes it well-suited for forecasting, classification, and anomaly detection tasks. Furthermore, QCAAPatchTF is a lightweight model that demonstrates state-of-the-art runtime efficiency compared to conventional approaches. Future work will focus on developing a quantum oracle to refine the attention mechanism, enhance computational efficiency, and explore its integration within large language models (LLMs) for time series analysis. In addition, optimizing quantum parameter tuning remains a key challenge in maximizing its effectiveness.

\section*{ACKNOWLEDGMENT}
This work is partially supported by the 'Resurssmarta Processor (RSP)', the Wallenberg AI, Autonomous Systems and Software Program (WASP), and the Wallenberg Initiative Materials Science for Sustainability (WISE), all funded by the Knut and Alice Wallenberg Foundation. 


\bibliographystyle{IEEEtran}
\bibliography{sample-base}

\begin{thebibliography}{10}
\providecommand{\url}[1]{#1}
\csname url@samestyle\endcsname
\providecommand{\newblock}{\relax}
\providecommand{\bibinfo}[2]{#2}
\providecommand{\BIBentrySTDinterwordspacing}{\spaceskip=0pt\relax}
\providecommand{\BIBentryALTinterwordstretchfactor}{4}
\providecommand{\BIBentryALTinterwordspacing}{\spaceskip=\fontdimen2\font plus
\BIBentryALTinterwordstretchfactor\fontdimen3\font minus \fontdimen4\font\relax}
\providecommand{\BIBforeignlanguage}[2]{{%
\expandafter\ifx\csname l@#1\endcsname\relax
\typeout{** WARNING: IEEEtran.bst: No hyphenation pattern has been}%
\typeout{** loaded for the language `#1'. Using the pattern for}%
\typeout{** the default language instead.}%
\else
\language=\csname l@#1\endcsname
\fi
#2}}
\providecommand{\BIBdecl}{\relax}
\BIBdecl

\bibitem{zhang2024hybrid}
J.~Zhang, H.~Liu, W.~Bai, and X.~Li, ``A hybrid approach of wavelet transform, arima and lstm model for the share price index futures forecasting,'' \emph{The North American Journal of Economics and Finance}, vol.~69, p. 102022, 2024.

\bibitem{shi2025digital}
S.~Shi, N.~Wang, S.~Chen, B.~Hu, J.~Peng, and Z.~Shi, ``Digital mapping of soil salinity with time-windows features optimization and ensemble learning model,'' \emph{Ecological Informatics}, vol.~85, p. 102982, 2025.

\bibitem{yaprakdal2023multivariate}
F.~Yaprakdal and M.~Varol~Ar{\i}soy, ``A multivariate time series analysis of electrical load forecasting based on a hybrid feature selection approach and explainable deep learning,'' \emph{Applied Sciences}, vol.~13, no.~23, p. 12946, 2023.

\bibitem{engel2024transformer}
E.~A. Engel and N.~E. Engel, ``A transformer with a fuzzy attention mechanism for weather time series forecasting,'' in \emph{International Conference on Neuroinformatics}.\hskip 1em plus 0.5em minus 0.4em\relax Springer, 2024, pp. 418--425.

\bibitem{kong2025deep}
X.~Kong, Z.~Chen, W.~Liu, K.~Ning, L.~Zhang, S.~Muhammad~Marier, Y.~Liu, Y.~Chen, and F.~Xia, ``Deep learning for time series forecasting: A survey,'' \emph{International Journal of Machine Learning and Cybernetics}, pp. 1--34, 2025.

\bibitem{xu2021anomaly}
J.~Xu, ``Anomaly transformer: Time series anomaly detection with association discrepancy,'' \emph{arXiv preprint arXiv:2110.02642}, 2021.

\bibitem{wen2022transformers}
Q.~Wen, T.~Zhou, C.~Zhang, W.~Chen, Z.~Ma, J.~Yan, and L.~Sun, ``Transformers in time series: A survey,'' \emph{arXiv preprint arXiv:2202.07125}, 2022.

\bibitem{morid2023time}
M.~A. Morid, O.~R.~L. Sheng, and J.~Dunbar, ``Time series prediction using deep learning methods in healthcare,'' \emph{ACM Transactions on Management Information Systems}, vol.~14, no.~1, pp. 1--29, 2023.

\bibitem{cajachagua2025impact}
K.~N. Cajachagua-Torres, M.~O. Xavier, H.~G. Quezada-Pinedo, C.~A. Huayanay-Espinoza, A.~G.~O. Rios, A.~Amouzou, A.~Ma{\"\i}ga, N.~Akseer, A.~Matijasevich, and L.~Huicho, ``Impact of the covid-19 pandemic on small vulnerable newborns: an interrupted time series analysis in peru and brazil,'' \emph{Journal of Global Health}, vol.~15, p. 04026, 2025.

\bibitem{zhang2024self}
K.~Zhang, Q.~Wen, C.~Zhang, R.~Cai, M.~Jin, Y.~Liu, J.~Y. Zhang, Y.~Liang, G.~Pang, D.~Song \emph{et~al.}, ``Self-supervised learning for time series analysis: Taxonomy, progress, and prospects,'' \emph{IEEE Transactions on Pattern Analysis and Machine Intelligence}, 2024.

\bibitem{li2023transferable}
Z.~Li, R.~Cai, T.~Z. Fu, Z.~Hao, and K.~Zhang, ``Transferable time-series forecasting under causal conditional shift,'' \emph{IEEE Transactions on Pattern Analysis and Machine Intelligence}, 2023.

\bibitem{spadon2021pay}
G.~Spadon, S.~Hong, B.~Brandoli, S.~Matwin, J.~F. Rodrigues-Jr, and J.~Sun, ``Pay attention to evolution: Time series forecasting with deep graph-evolution learning,'' \emph{IEEE Transactions on Pattern Analysis and Machine Intelligence}, vol.~44, no.~9, pp. 5368--5384, 2021.

\bibitem{biamonte2017quantum}
J.~Biamonte, P.~Wittek, N.~Pancotti, P.~Rebentrost, N.~Wiebe, and S.~Lloyd, ``Quantum machine learning,'' \emph{Nature}, vol. 549, no. 7671, pp. 195--202, 2017.

\bibitem{cerezo2022challenges}
M.~Cerezo, G.~Verdon, H.-Y. Huang, L.~Cincio, and P.~J. Coles, ``Challenges and opportunities in quantum machine learning,'' \emph{Nature Computational Science}, vol.~2, no.~9, pp. 567--576, 2022.

\bibitem{peral2024systematic}
D.~Peral-Garc{\'\i}a, J.~Cruz-Benito, and F.~J. Garc{\'\i}a-Pe{\~n}alvo, ``Systematic literature review: Quantum machine learning and its applications,'' \emph{Computer Science Review}, vol.~51, p. 100619, 2024.

\bibitem{don2024fusion}
A.~K.~K. Don, I.~Khalil, and M.~Atiquzzaman, ``A fusion of supervised contrastive learning and variational quantum classifiers,'' \emph{IEEE Transactions on Consumer Electronics}, 2024.

\bibitem{gohel2024quantum}
P.~Gohel and M.~Joshi, ``Quantum time series forecasting,'' in \emph{Sixteenth International Conference on Machine Vision (ICMV 2023)}, vol. 13072.\hskip 1em plus 0.5em minus 0.4em\relax SPIE, 2024, pp. 390--398.

\bibitem{shi2024qsan}
J.~Shi, R.-X. Zhao, W.~Wang, S.~Zhang, and X.~Li, ``Qsan: A near-term achievable quantum self-attention network,'' \emph{IEEE Transactions on Neural Networks and Learning Systems}, 2024.

\bibitem{wu2023quantum}
H.~Wu, J.~Zhou, Q.~Zhang, Y.~Lei, K.~Yu, W.~An, and J.~Zhang, ``A quantum-based attention mechanism in scene text detection,'' in \emph{Chinese Conference on Pattern Recognition and Computer Vision (PRCV)}.\hskip 1em plus 0.5em minus 0.4em\relax Springer, 2023, pp. 3--14.

\bibitem{hsu2024quantum}
Y.-C. Hsu, N.-Y. Chen, T.-Y. Li, K.-C. Chen \emph{et~al.}, ``Quantum kernel-based long short-term memory for climate time-series forecasting,'' \emph{arXiv preprint arXiv:2412.08851}, 2024.

\bibitem{thakkar2024improved}
S.~Thakkar, S.~Kazdaghli, N.~Mathur, I.~Kerenidis, A.~J. Ferreira-Martins, and S.~Brito, ``Improved financial forecasting via quantum machine learning,'' \emph{Quantum Machine Intelligence}, vol.~6, no.~1, p.~27, 2024.

\bibitem{zhao2024qksan}
R.-X. Zhao, J.~Shi, and X.~Li, ``Qksan: A quantum kernel self-attention network,'' \emph{IEEE Transactions on Pattern Analysis and Machine Intelligence}, 2024.

\bibitem{vaswani2017attention}
A.~Vaswani, ``Attention is all you need,'' \emph{Advances in Neural Information Processing Systems}, 2017.

\bibitem{nie2022time}
Y.~Nie, N.~H. Nguyen, P.~Sinthong, and J.~Kalagnanam, ``A time series is worth 64 words: Long-term forecasting with transformers,'' \emph{arXiv preprint arXiv:2211.14730}, 2022.

\bibitem{kitaev2020reformer}
N.~Kitaev, {\L}.~Kaiser, and A.~Levskaya, ``Reformer: The efficient transformer,'' \emph{arXiv preprint arXiv:2001.04451}, 2020.

\bibitem{zeng2023transformers}
A.~Zeng, M.~Chen, L.~Zhang, and Q.~Xu, ``Are transformers effective for time series forecasting?'' in \emph{Proceedings of the AAAI conference on artificial intelligence}, vol.~37, no.~9, 2023, pp. 11\,121--11\,128.

\bibitem{zerveas2021transformer}
G.~Zerveas, S.~Jayaraman, D.~Patel, A.~Bhamidipaty, and C.~Eickhoff, ``A transformer-based framework for multivariate time series representation learning,'' in \emph{Proceedings of the 27th ACM SIGKDD conference on knowledge discovery \& data mining}, 2021, pp. 2114--2124.

\bibitem{zhou2021informer}
H.~Zhou, S.~Zhang, J.~Peng, S.~Zhang, J.~Li, H.~Xiong, and W.~Zhang, ``Informer: Beyond efficient transformer for long sequence time-series forecasting,'' in \emph{Proceedings of the AAAI conference on artificial intelligence}, vol.~35, no.~12, 2021, pp. 11\,106--11\,115.

\bibitem{chen2021autoformer}
M.~Chen, H.~Peng, J.~Fu, and H.~Ling, ``Autoformer: Searching transformers for visual recognition,'' in \emph{Proceedings of the IEEE/CVF international conference on computer vision}, 2021, pp. 12\,270--12\,280.

\bibitem{liu2023itransformer}
Y.~Liu, T.~Hu, H.~Zhang, H.~Wu, S.~Wang, L.~Ma, and M.~Long, ``itransformer: Inverted transformers are effective for time series forecasting,'' \emph{arXiv preprint arXiv:2310.06625}, 2023.

\bibitem{zhou2022fedformer}
T.~Zhou, Z.~Ma, Q.~Wen, X.~Wang, L.~Sun, and R.~Jin, ``Fedformer: Frequency enhanced decomposed transformer for long-term series forecasting,'' in \emph{International conference on machine learning}.\hskip 1em plus 0.5em minus 0.4em\relax PMLR, 2022, pp. 27\,268--27\,286.

\bibitem{woo2022etsformer}
G.~Woo, C.~Liu, D.~Sahoo, A.~Kumar, and S.~Hoi, ``Etsformer: Exponential smoothing transformers for time-series forecasting,'' \emph{arXiv preprint arXiv:2202.01381}, 2022.

\bibitem{chakraborty2024edformer}
S.~Chakraborty, I.~Delibasoglu, and F.~Heintz, ``Edformer: Embedded decomposition transformer for interpretable multivariate time series predictions,'' \emph{arXiv preprint arXiv:2412.12227}, 2024.

\bibitem{zhang2023crossformer}
Y.~Zhang and J.~Yan, ``Crossformer: Transformer utilizing cross-dimension dependency for multivariate time series forecasting,'' in \emph{The eleventh international conference on learning representations}, 2023.

\bibitem{zhang2025diffusion}
S.~Zhang, Z.~Qin, Y.~Zhang, Y.~Zhou, R.~Li, C.~Du, and Z.~Xiao, ``Diffusion-enhanced optimization of variational quantum eigensolver for general hamiltonians,'' \emph{arXiv preprint arXiv:2501.05666}, 2025.

\bibitem{maheshwari2021variational}
D.~Maheshwari, D.~Sierra-Sosa, and B.~Garcia-Zapirain, ``Variational quantum classifier for binary classification: Real vs synthetic dataset,'' \emph{IEEE access}, vol.~10, pp. 3705--3715, 2021.

\bibitem{zhou2023multi}
J.~Zhou, D.~Li, Y.~Tan, X.~Yang, Y.~Zheng, and X.~Liu, ``A multi-classification classifier based on variational quantum computation,'' \emph{Quantum Information Processing}, vol.~22, no.~11, p. 412, 2023.

\bibitem{campagner2024ensemble}
A.~Campagner, M.~Barandas, D.~Folgado, H.~Gamboa, and F.~Cabitza, ``Ensemble predictors: Possibilistic combination of conformal predictors for multivariate time series classification,'' \emph{IEEE Transactions on Pattern Analysis and Machine Intelligence}, 2024.

\bibitem{sa_timeseries}
\BIBentryALTinterwordspacing
UVA-MLSys, ``Sa-timeseries: Self-attention time series models,'' 2024, last accessed: 10 March 2025. [Online]. Available: \url{https://github.com/UVA-MLSys/SA-Timeseries}
\BIBentrySTDinterwordspacing

\bibitem{pems_dataset}
\BIBentryALTinterwordspacing
E.~Mahy, ``Pems dataset,'' 2024, last accessed: 10 March 2025. [Online]. Available: \url{https://www.kaggle.com/datasets/elmahy/pems-dataset}
\BIBentrySTDinterwordspacing

\bibitem{bagnall2018uea}
A.~Bagnall, H.~A. Dau, J.~Lines, M.~Flynn, J.~Large, A.~Bostrom, P.~Southam, and E.~Keogh, ``The uea multivariate time series classification archive, 2018,'' \emph{arXiv preprint arXiv:1811.00075}, 2018.

\end{thebibliography}

\vskip 0pt plus -1fil

\end{document}